# AI Playing Business Games: Benchmarking Large Language Models on Managerial Decision-Making in Dynamic Simulations

Berdymyrat Ovezmyradov[1]

Transport and Telecommunication Institute

Lauvas iela 2, Riga, LV 1019, Latvia

*berdyovezmuradov@gmail.com*

## Abstract

The rapid advancement of LLMs sparked significant interest in their potential to augment or automate managerial functions. One of the most recent trends in AI benchmarking is performance of Large Language Models (LLMs) over longer time horizons. While LLMs excel at tasks involving natural language and pattern recognition, their capabilities in multi-step, strategic business decision-making remain largely unexplored. Few studies demonstrated how results can be different from benchmarks in short-term tasks. Meanwhile, there is a shortage of alternative benchmarks for long-term coherence. This research analyses a novel benchmark using a business game for the decision making in business. The research contributes to the recent literature on AI by proposing a reproducible, open-access management simulator to the research community for LLM benchmarking. This novel framework offers the first spreadsheet-based benchmark for LLM managerial decision-making over extended period of one year of a retail operation. It is used for evaluating the performance of leading LLMs available in free online interface: Gemini, ChatGPT, Meta AI, Mistral AI, and Grok. LLM makes decisions for a simulated retail company. A dynamic, month-by-month management simulation provides transparently in spreadsheet model as experimental environment. In each of twelve months, the LLMs are provided with a structured prompt containing a full business report from the previous period and are tasked with making key strategic decisions: pricing, order size, marketing budget, hiring, dismissal, loans, training expense, R&D expense, sales forecast, income forecast The methodology is designed to compare the LLMs on quantitative metrics: profit, revenue, and market share, and other KPIs. LLM decisions are also analyzed in their strategic coherence, adaptability to market changes, and the rationale provided for their decisions. This approach allows to move beyond simple performance metrics for assessment of the long-term decision-making in business based on a comprehensive set of historical indicators. The provision of the open-access simulator file enables benchmarking for future research in AI business decision-making. The findings offer insights into the strengths and limitations of current LLMs in a managerial context, informing both the development of AI-driven decision support systems and the future of business education. Pivots to AI in multi-step management simulation, a task largely absent from current benchmarks, allows systematic comparison of different LLMs.

[1] Corresponding author

## 1 Introduction

Managerial decision-making is a complex process that extends beyond simple data analysis. It involves synthesizing financial reports, market conditions, competitor actions, and multiple goals to make integrated decisions across multiple functional areas, such as procurement, pricing, and marketing. Management simulations have long served as a tool for both research and education to model this complexity. By immersing participants in a dynamic, competitive environment, these simulations provide a controlled setting to test. The intersection of generative AI and management simulation therefore presents a novel opportunity to evaluate the capabilities of LLMs in a realistic, albeit virtual, environment.

While prior research has explored the use of artificial intelligence (AI) in management simulations, much of it has focused on specific applications. The core of this literature has largely centered on narrowly focused decisions for a particular function. With the advent of multi-purpose LLMs, there is a distinct need to benchmark these models' performance on the integrated and high-level decision-making required to run an entire business for multiple periods. This study addresses this research gap by introducing a novel, reproducible framework to evaluate and compare the decision-making of leading LLMs in a dynamic management simulation (hereinafter, the simulation)

.AI has been employed for tasks such as data analysis, predictive modeling, and process automation, primarily as a decision support tool. However, the emergence of advanced models like ChatGPT and Gemini marks a paradigm shift. These generative systems possess sophisticated capabilities that enable them to generate creative solutions and engage in multi-step problem-solving. Vending-Bench recently demonstrated how all large language models (LLMs) struggle in business management with long-horizon coherence (Backlund and Petersson, 2025). The integration of AI into business operations might fundamentally reshape organizational dynamics and competitive landscapes in the future, but their current state does not allow reliable decision making. This raises a timely research question: to what extent can LLMs perform the holistic and adaptive decision-making to manage a business?

The study aims to provide a comparison of the performance of LLMs freely accessible to users against the same set of conditions over a twelve-month period. These LLMs in the study represent both commercial and open-source alternatives: Gemini, ChatGPT, Meta AI, Mistral AI, and Grok, We have developed a comprehensive spreadsheet model in an accessible, open-source Excel file, allowing for reproducibility. This approach allows to compare LLM decision-making, in terms of coherence and adaptability in a more realistic simulation across an extended range of metrics. These are summarized in separate figures and tables in the file. The study also explores responses for understanding models' foresight, risk assessment, and justification of decisions. The simulation provides a risk-free environment for testing hypotheses, developing practical skills, and understanding the complex, interconnected functions of a business. That was an original purpose of the simulator. But it can also be used for AI benchmarking, as this study demonstrates. This simulation can be downloaded here: *https://sourceforge.net/projects/supply-chain-competition-game/files/.*

The study contributes to the rapidly expanding literature on AI by providing a benchmark in a wider range of the financial outcomes and other business metrics of firms established in OR/MS" (Operations Research/Management Science) literature. The benchmark is novel since it tests capability of LLM to make tactical decisions based on a wider range of common financial and other indicators (in contrast, Vending-Bench analyzed performance in daily purchasing, hiring, communication, and other operational decisions). This analysis can provide deeper insights behind the quantitative results.

The next sections present the literature review, methods, results, and summary.

## 2 Review of literature on serious games and AI

This section covers two streams of business literature relevant to the study: AI benchmarking and serious games.

Early research on large language models (LLMs) related to business focused on AI as a support tool to test its accuracy in general fields and tasks on only a narrow subset of datasets (Wang et al. 2018; Srivastava et al. 2023; Liu et al. 2023; Wang et al. 2024; Joshi 2025; Cooper et al. 2025; Kazemi et al. 2025). This body of work demonstrated LLM's capacity in certain areas of knowledge to process vast datasets matching human capability, while potentially leading to improved economic processes (though still remaining limited in many fields). Recent advancements have fundamentally altered the landscape of AI capabilities, prompting a new wave of inquiry into their potential as autonomous agents in specialized managerial tasks such as finance, logistics and marketing (Mikhaylov 2021; Xie et al. 2023; Shabsigh and Boukherouaa 2023; Xie et al. 2024; Özgül and Kahraman 2025; Jin et al. 2024; Srivastava et al. 2024; Mohsin 2025; Mattusch 2025; Liu et al. 2025).Research has begun to explore the use of LLMs in simulations, to show whether they could have competence in rational decision-making performing tasks requiring a high degree of understanding the complex interaction effects between various factors in OR/MS (Kurter 2025; Cardoso et al. 2023; Csaszar et al.2024). Notably, Vending-Bench, a simulated environment to test ability operating a vending machine, shows how LLMs might be with coherence in long-term decision making (Backlund and Petersson, 2025). The integration of AI with the advent of sophisticated generative models into managerial decision-making remains a nascent concept. Unlike earlier AI models, LLMs are not trained for a single, narrow task. Instead, their pre-training on vast datasets has endowed them with a remarkable capacity for general-purpose reasoning and the ability to maintain context over extended conversations.

The use of simulation games has been a parallel and long-standing area of research in both management and education, but not so much in AI. Visibility, reproducibility, safety, economy, and availability of computer-based serious games present a viable alternative for training on the job, providing support for operations managers in learning decision-making while experimenting within a safe learning environment (Strozzi et al. 2007). Such games were instrumental in industry simulation for developing business strategy models and facilitating executive debate (Ninios et al. 1995, Ponte et al. 2016). Management is a key word in business education (Park 2021). Supply chain management and logistics is one of its most important fields. Modeling supply chains together with manufacturing and service quality was among the most widespread OR/MS games categories, with the beer game (also known as beer distribution game) ranked at the top of the related content (Lewis and Maylor 2007). MIT first introduced the original beer distribution game in the early 1960s, and it soon became a popular tool for explaining the dynamics of the supply chain (Sterman 1989). The bullwhip effect represents one of the experimental results of the beer game (Lee and Padmanabhan 1997). The beer game really represents a classic system dynamics model used not only exclusively for educational purposes but also decision-making in real business cases and research. Besides the beer game, Goldratt's Game is another general class of exercises to help understand the operations with queues or inventory buffers in manufacturing, for which Excel-based games were developed (Johnson and Drougas 2002). The Logistics Game is not the first free enhanced version of the beer game. For instance, MIT (2022) is the birth place of the beer game that offers a free online option for using it for instruction. This and many other providers of management simulation tools chose a web interface, which understandably restricts the level of the game design modification according to an instructor's needs. Furthermore, the necessity of internet connection and registration (often with the verification of being an educator at an academic institution) somewhat restricts the availability of such resources. Consequently, opting for electronic spreadsheet versions instead of web-based format should be mentioned. Spreadsheets have long been used as a tool for serious or simulation games, allowing easier tracking of the progress of students (Strakos 2016, Fetter and Shockley 2014). Spreadsheets help prepare future managers in OR/MS education (Albright and Winston 2005, Gardner 2008). Practitioners using spreadsheets often positively evaluate tools that facilitate "what-if" analyses (Althuizen et al. 2012). However, the role of automation within these simulations has largely been confined to creating rule-based competitors or providing automated feedback, rather than serving as a high-level, strategic agent itself. The beer game can even be helpful in applications of promising AI techniques in logistics. Artificial agents managed by computer programs effectively adapted order policies in the game with supply chains, intractable when using analytic methods (Kimbrough 2002). Genetic algorithms found an optimal policy in the game when customers' demand increased (Strozzi 2007). Reinforcement learning allowed for finding good policies where analytical solutions were unavailable (Chaharsooghi et al. 2008). AI outperformed humans in certain tasks when playing a popular team-based game on the Nintendo Switch with players acting as workers hired by firms, but the automation could reduce overall performance, team trust, and individual effort (Dell'Acqua et al 2025). Despite few papers involving AI in business games, the related research still has focused more on how computers assist human players rather than how it performs in an autonomous managerial role. In the primary motivation and development behind this work, it is believed that the management simulation would offer more freedom and empowerment not only in using management simulators for education but also AI benchmarking.

The two streams of literature above suggest that LLMs in management simulations can act as effective agents within complex systems, capable of making decisions that are not explicitly programmed but are inferred from a broad base of knowledge. Despite these promising developments, a significant research gap remains. There is a lack of comparative studies that benchmark different LLMs on a holistic, multi-turn decision-making like managing a retail business as a virtual CEO. Existing AI benchmarks primarily focus on discrete tasks like code generation or question-answering, which fail to capture the dynamic, integrated nature of a competitive business environment, as Vending-Bench revealed. Our proposed research directly addresses this gap by creating a novel benchmark that leverages a dynamic management simulation to systematically compare the performance of leading LLMs. Unlike

Vending-Bench, our benchmark does not focus on daily operations. Instead, we propose benchmarking performance in tactical decision-making using standard financial statements and a set of KPIs and historical data applicable to a wider range of real business scenarios. This approach will not only contribute to the understanding of LLM capabilities in a managerial context but also provide a reproducible framework for future research in AI business strategy, building upon the foundational work in both AI decision support systems and management simulation.

## 3 Methods

This study employs a spreadsheet model of a management simulator to evaluate the decision-making capabilities of leading large language models (LLMs) in a business environment. The methodology is structured to ensure transparency and reproducibility for comparison.

The core research instrument is a comprehensive, open-source management simulation developed in Microsoft Excel. The simulator models the operations of a fictional manufacturing company, RETAILER ONE competing in a dynamic market with a single-product over a 12-month period. The simulation includes interdependent variables that react to decisions, creating a realistic, closed-loop environment. Key decision variables for each month are: (1) product price, (2) production order size, and (3) marketing budget; (4) workers dismissed/hired; (5) dividends %; (6) loans; (7) training expense; (8) R&D expense; (9) sales forecast next period; (10) net income forecast. More details about the simulator were published by Ovezmyradov et al. (2026).

Performance metrics and other outputs, which serve as the basis for the next round's decisions, include sales revenue, cost of goods sold (COGS), total profit/loss, ending cash on hand, inventory levels, and numerous other indicators outlined in Tables 1, 2, and 3. The simulation incorporates dynamic elements such as pricing and a market demand function that is sensitive to pricing, marketing, and competitor actions.

The Excel file is fully flexible to adjust with formulas and open access ensures that any researcher can replicate the simulation to verify the results. This file is made available in a public repository together with a guide and other supplementary resources to start using simulator: *https://sourceforge.net/projects/supply-chain-competition-game/files/* (file of the simulator is “OnePlayer2024r1protectedblank8.xlsx”). Appendices to this paper show details of the simulation and corresponding prompts.

The study's comparators to test the benchmark are the following leading LLMs accessible from online interface on website as of the mid-2025:

1. ChatGPT-5: The most popular LLM by OpenAI, recognized for its strong performance on a wide range of tasks (*chatgpt.com*).
2. Gemini 2.5 Flash: A multimodal model by Google with longer context window relative to other free alternatives (*gemini.google.com*).
3. Gemini 2.5 Pro: The only advanced reasoning model in the comparisons available for free by Google (though with restrictive limits – the same link as above).
4. Grok: A newer model from xAI (*grok.com*).
5. Meta AI: The most popular and widely-used open-source LLM (with the currently used generation being Meta Llama 4 versions) by Meta (*meta.ai*).
6. Mistral AI: A leading LLM (with several currently used versions of Mistral) by a European AI company (*chat.mistral.ai*).

There are other LLMs accessible in commercial and open-source forms, but the six alternatives above were considered relevant in terms of being safely accessible online. Each LLM was treated in independent experimental conditions, with a dedicated session to ensure that no context influenced it. The procedure was as follows.

Initial Prompting (Month 1) shown in Appendix A: The initial prompt, identical for each LLM, established the role of the AI as the CEO of the retailer. It provided a detailed report of the previous fiscal year, the company's starting position, and the market conditions for start of January. The prompt concluded with a request for the key decisions for January in a clear, structured format.

Sequential Decision-Making (Months 2-12) presented by Appendix B: For each subsequent month, the procedure was repeated with another prompt followed by the past result(s). The simulated results from the previous month (e.g., January's sales, profit, and ending cash) were copied from the Excel file and incorporated into a new, updated prompt for the next month.

This iterative process was carried out for all 12 months, creating a continuous, dynamic feedback loop for each AI. The prompt structure and language were kept consistent (with only minor modification across sessions when appropriate due to interface) across all months and all LLMs to minimize the influence of variation. All decisions made by the LLMs and the corresponding simulated outcomes were recorded. The rationale and qualitative commentary provided by the LLMs for their decisions were also saved on a corresponding site session for subsequent qualitative analysis.

The collected data was used for quantitative analysis with visualizations of the two most important metrics: sales revenue and net income. We also compare the performance of the LLM-driven companies using a variety of additional metrics, including market share. The percentage of the total market that each company captured, for instance, provide insight into their competitive positioning. There are other metrics outlined in tables. We use descriptive statistics to summarize the performance determine differences between the different LLMs.

The data (the LLMs' rationales for their decisions) is also analyzed to assess the coherence and adaptability of each model. The simulation is designed with dynamic elements, such as changes in market demand. We analyze how each LLM reacts to these changes. A model demonstrating strong adaptability would adjust its pricing and production to mitigate losses or capitalize on new opportunities, while a less adaptive model might continue with a sub-optimal strategy. The qualitative rationales provided by the LLMs can be crucial here, as they indicate whether the AI recognized and responded to the market shock. We conduct a content analysis of the textual justifications provided by each LLM for its monthly decisions. We imply the following aspects: data-driven decisions explicitly linked to previous month's metrics, foresight with decisions that anticipate future market conditions; risk assessment with rationales that acknowledge potential risks (e.g., overproduction, pricing), strategic language in the use of sophisticated business terminology and concepts.

We compare the different key performance indicators (KPIs) for each LLM-managed company, but focus on income statement. The total revenue and profit or loss accrued over the 12-month period. They serve as the primary metric for overall success. A month-by-month analysis of revenue and profit to identify trends, volatility, and the impact of specific decisions. To take into account both sales and profits, a combined metric was used for identification of the best performing comparators - ratio of profit to sales as in Eq. 1:

$$Net\ Profit\ Margin = \frac{Net\ Income}{Sales\ Revenue}. \quad (1)$$

Data for sales and profits is visualized using time-series graphs to illustrate the monthly progression of key metrics like profit and inventory levels for each LLM. This provides a clear visual to compare their performance.

Data driven decisions in management science requires knowledge of complex OR/MS models (Hillier and Lieberman, 2004; Albright and Winston, 2005).We look for patterns that align with established models (e.g., Ansoff, 1965; Chandler, 1962). For example, a model that consistently invests in marketing to support a high-price or other consistent strategy depending on the market conditions could be considered as "coherent," whereas a model with erratic month-to-month changes in price and orders would be regarded as "incoherent." The further analysis is designed to compare numerical output first and then move beyond the numbers to provide qualitative insight into the possible thinking, or lack thereof, behind the LLMs' decisions. Thus, the methodology utilizing a spreadsheet model provides a novel framework for benchmarking LLMs on a complex, long-term, and multi-faceted task, contributing a new alternative for evaluating AI performance in real-world business scenarios.

We acknowledge limitation of the chosen methods, the main of which is the limited number of experiments and LLM models due to time and space constraints. Some of the newer and increasingly popular services (notably, recent powerful Chinese LLMs) are therefore excluded. No statistical tests are conducted due to the same reason. Furthermore, LLMs might have different levels of sensitivity and, potentially, misinterpretation to prompt phrasing. The main purpose is to propose a new benchmark with only exploratory comparisons. More comprehensive results will hopefully follow. The last section briefly discusses these as future work directions.

## 4 Results

Figure 1 describes overall performance of the comparators for all periods.

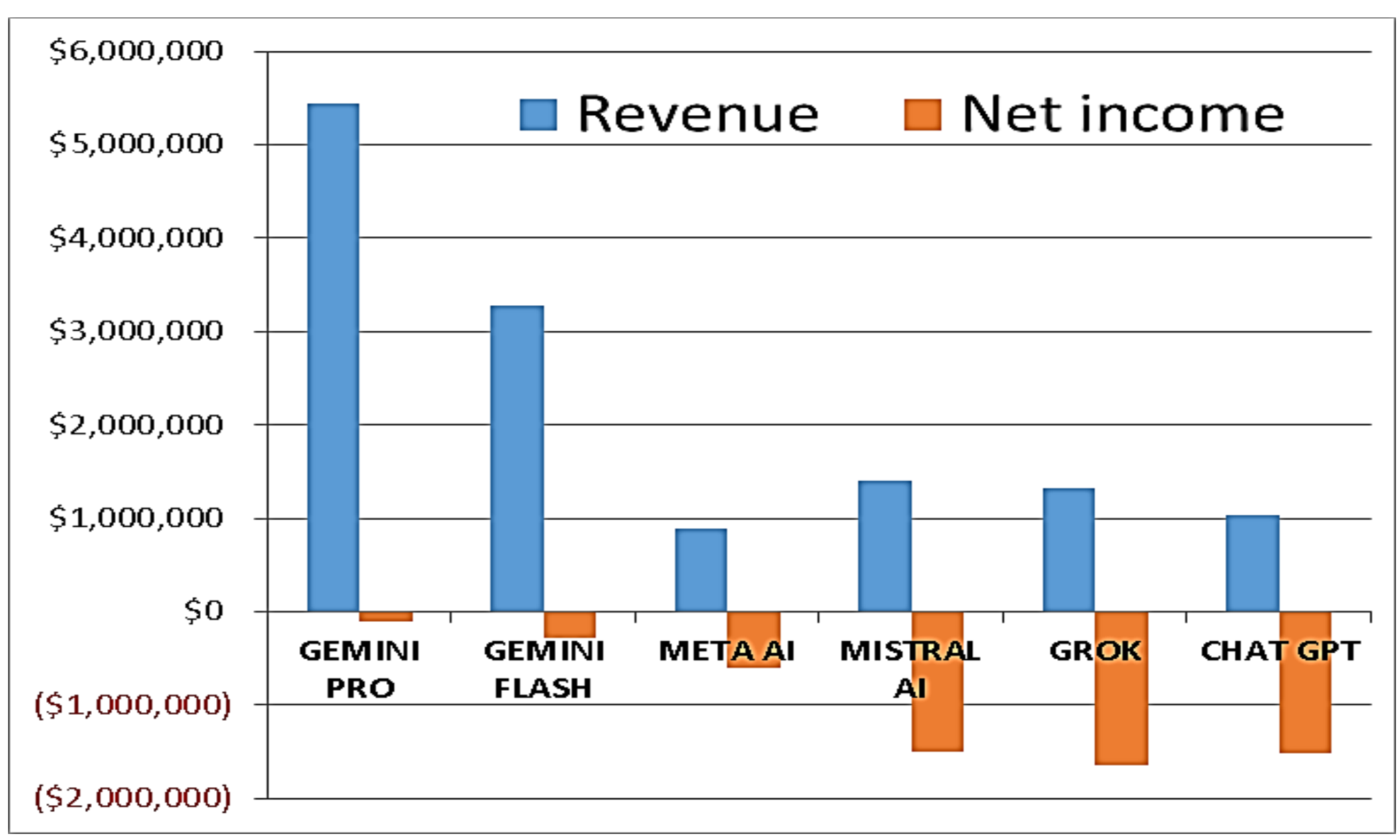

*Fig. 1 Performance across comparator LLMs.*

Tables 1, 2 and 3 provide comparative statics across all the indicators in the simulator output. These provide the context to the quantitative data outside the figures, revealing the strengths and weaknesses of each LLM as a strategic agent and providing a richer understanding of their capabilities in diverse metrics.

**Table 1** Financial statements*

| | **MISTRAL** | **GEMINI FLASH** | **META** | **GROK** | **CHAT GPT** | **GEMINI PRO** |
|---|---|---|---|---|---|---|
| INCOME STATEMENT | | | | | | |
| REVENUE | $1,396,902 | $3,283,242 | $881,049 | $1,319,555 | $1,040,089 | $5,444,246 |
| Materials expense | $989,405 | $2,308,244 | $578,727 | $878,347 | $758,054 | $4,112,558 |
| Staff costs | $374,057 | $369,823 | $209,032 | $290,396 | $387,180 | $373,357 |
| Depreciation expense | $84,000 | $84,000 | $84,000 | $84,000 | $84,000 | $84,000 |
| Other operating expenses | $1,325,122 | $689,659 | $526,976 | $1,637,525 | $1,267,354 | $765,964 |
| TOTAL COSTS AND EXPENSES | $2,772,584 | $3,451,726 | $1,398,735 | $2,890,268 | $2,496,587 | $5,335,879 |
| OPERATING INCOME | ($1,375,682) | ($168,484) | ($517,686) | ($1,570,713) | ($1,456,498) | $108,367 |
| Interest expense | $120,000 | $60,000 | $67,500 | $60,000 | $60,000 | $165,000 |
| PROFIT BEFORE TAX | ($1,495,682) | ($228,484) | ($585,186) | ($1,630,713) | ($1,516,498) | ($56,633) |
| Income tax expense | $7,873 | $45,613 | $15,177 | $7,831 | $ - | $47,546 |
| NET INCOME | ($1,503,555) | ($274,097) | ($600,363) | ($1,638,544) | ($1,516,498) | ($104,179) |
| BALANCE SHEET | | | | | | |

| ASSETS | | | | | | |
|---|---|---|---|---|---|---|
| Cash (overdraft if negative) | 835,359 | 1,025,769 | 1,088,102 | 720,646 | 607,237 | 543,568 |
| Accounts receivable | 116,408 | 273,604 | 73,421 | 109,963 | 86,674 | 453,687 |
| Inventory | 3,920 | 67,951 | 4,270 | 11,922 | 629 | 397,773 |
| Total current assets | 955,687 | 1,367,324 | 1,165,793 | 842,531 | 694,540 | 1,395,028 |
| Non-current assets | | | | | | |
| Buildings | 1,000,000 | 1,000,000 | 1,000,000 | 1,000,000 | 1,000,000 | 1,000,000 |
| Accumulated depreciation | 13,000 | 13,000 | 13,000 | 13,000 | 13,000 | 13,000 |
| Equipment | 500,000 | 500,000 | 500,000 | 500,000 | 500,000 | 500,000 |
| Accumulated depreciation | 32,500 | 32,500 | 32,500 | 32,500 | 32,500 | 32,500 |
| Intangible assets | 100,000 | 100,000 | 100,000 | 100,000 | 100,000 | 100,000 |
| Total non-current assets | 1,554,500 | 1,554,500 | 1,554,500 | 1,554,500 | 1,554,500 | 1,554,500 |
| TOTAL ASSETS | 2,510,187 | 2,921,824 | 2,720,293 | 2,397,031 | 2,249,040 | 2,949,528 |
| EQUITY AND LIABILITIES | | | | | | |
| Accounts payable | 2,000 | 2,000 | 2,000 | 2,000 | 2,000 | 2,000 |
| Total current liabilities | 2,000 | 2,000 | 2,000 | 2,000 | 2,000 | 2,000 |
| Non-current liabilities | | | | | | |
| Long-term debt | 200,000 | 100,000 | 112,500 | 100,000 | 100,000 | 275,000 |
| Provisions | 8,211 | 8,789 | 5,664 | 7,009 | 8,777 | 9,298 |
| Total non-current liabilities | 208,211 | 108,789 | 118,164 | 107,009 | 108,777 | 284,298 |
| Equity | | | | | | |
| Paid-in capital | 2,848,000 | 2,848,000 | 2,848,000 | 2,848,000 | 2,848,000 | 2,848,000 |
| Retained earnings | -548,024 | -36,966 | -247,871 | -559,978 | -709,738 | -184,770 |
| Total equity | 2,299,976 | 2,811,034 | 2,600,129 | 2,288,022 | 2,138,262 | 2,663,230 |
| TOTAL EQUITY AND LIABILITIES | 2,510,187 | 2,921,824 | 2,720,293 | 2,397,031 | 2,249,040 | 2,949,528 |
| STATEMENT OF CASH FLOW | | | | | | |
| Net income | -125,296 | -22,841 | -50,030 | -136,545 | -126,375 | -8,682 |
| Depreciation and amortization | 7,000 | 7,000 | 7,000 | 7,000 | 7,000 | 7,000 |
| Changes in inventory | -29,167 | -29,167 | -29,167 | -29,167 | -29,167 | -29,167 |
| Changes in provisions | 1,181 | 1,222 | 642 | 856 | 1,275 | 1,178 |
| Changes in receivables | - | - | - | - | - | 65,132 |
| Loans | 45,833 | - | 4,167 | - | - | 8,333 |
| Net increase (decrease) in cash and cash equivalents | -42,115 | 14,547 | -9,054 | -100,567 | -88,933 | -32,122 |
| Cash and cash equivalents at beginning of period | 877,474 | 1,011,222 | 1,097,157 | 821,213 | 696,170 | 575,690 |
| Cash and cash equivalents at end of period | 835,359 | 1,025,769 | 1,088,102 | 720,646 | 607,237 | 543,568 |

**For twelve months in a simulation session.*

**Table 2** Financial and forecasting results

| | MISTRAL | GEMINI FLASH | META | GROK | CHAT GPT | GEMINI PRO |
|---|---|---|---|---|---|---|

| ROA% | -5.4 | -0.81 | -1.88 | -6.76 | -5.98 | -0.3 |
|---|---|---|---|---|---|---|
| Leverage % | 10.99 | 3.93 | 4.65 | 4.91 | 5.37 | 10.84 |
| Gross profit margin (%) | -9% | 2% | -7% | 4% | -68% | -14% |
| Share price | 50 | 80 | 40 | 67 | 757 | 117 |
| Market capitalization | 1,420,510 | 2,286,898 | 1,130,637 | 1,915,515 | 21,570,965 | 3,330,795 |
| Sales forecast error % (MAPE) | 121 | 82 | 114 | 88 | 89 | 122 |
| Profit forecast error % (MAPE) | 108 | 117 | 150 | 117 | 116 | 411 |
| Market share % | 16% | 34% | 11% | 16% | 10% | 48% |

**Table 3** Human resources, environmental, and logistics performance.

| | MISTRAL | GEMINI FLASH | META | GROK | CHAT GPT | GEMINI PRO |
|---|---|---|---|---|---|---|
| Hiring | 13 | 9 | 4 | 8 | 10 | 9 |
| Redundancy | 2 | 0 | 4 | 7 | 0 | 4 |
| Hiring and dismissal cost | 34,000 | 18,000 | 24,000 | 44,000 | 20,000 | 34,000 |
| Worker wages | 283,381 | 293,186 | 154,193 | 205,330 | 305,983 | 282,797 |
| Sales and administrative wages | 56,676 | 58,637 | 30,839 | 41,066 | 61,197 | 56,559 |
| Training expense $ | 75,000 | 12,000 | 31,500 | 60,000 | 85,500 | 47,500 |
| Productivity ( hourly per worker) | 28 | 10 | 24 | 41 | 51 | 34 |
| Capacity utilization % | 5 | 15 | 3 | 3 | 2 | 8 |
| Average physical inventory | 1,360 | 3,679 | 908 | 1,343 | 1,059 | 9,798 |
| Inventory service level % (Fill Rate) | 44 | 84 | 33 | 44 | 31 | 96 |

For each of the 12 months, we recorded the decisions and the corresponding performance metrics from the simulations as outlined by Figures 2 to 7.

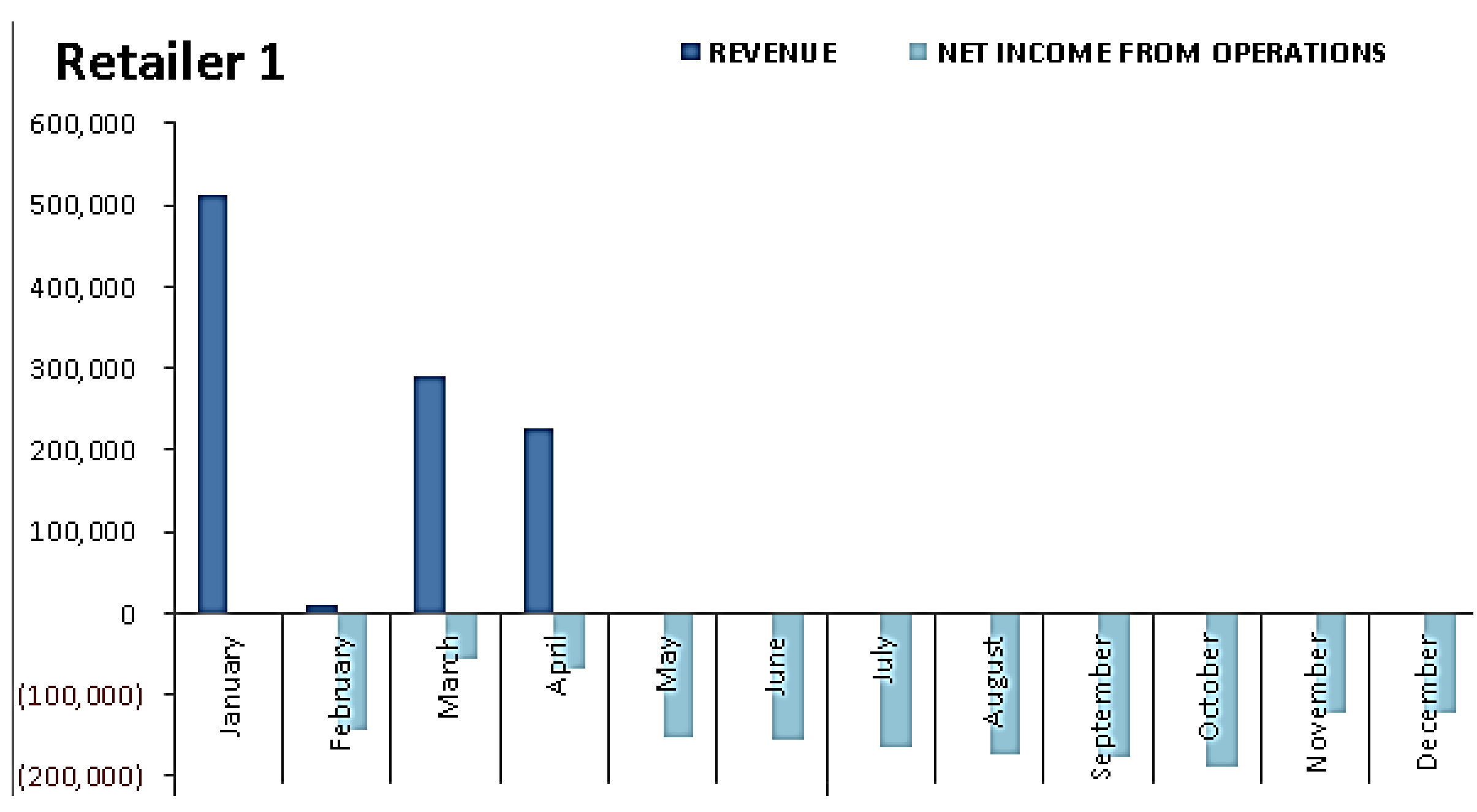

| Order in units | Price $ | Workers hired | Workers dismissed | Marketing expense $ | Loan $ | Training expense $ | R&D expense $ | Sales forecast next period $ | Net income forecast $ |
|---|---|---|---|---|---|---|---|---|---|
| 5000 | 105 | 2 | 0 | 20000 | 0 | 10000 | 10000 | 8000 | 25000 |
| 6000 | 102 | 0 | 0 | 25000 | 0 | 8000 | 8000 | 6000 | 30000 |
| 3000 | 95 | 0 | 0 | 15000 | 0 | 5000 | 0 | 120000 | -20000 |
| 6000 | 110 | 0 | 0 | 25000 | 0 | 6000 | 8000 | 6500 | 50000 |
| 7000 | 115 | 1 | 0 | 28000 | 0 | 7000 | 9000 | 7500 | 55000 |
| 6500 | 120 | 1 | 0 | 30000 | 0 | 7000 | 10000 | 8000 | 60000 |
| 7000 | 125 | 1 | 0 | 35000 | 0 | 7000 | 12000 | 10000 | 65000 |
| 8000 | 130 | 1 | 0 | 38000 | 0 | 7000 | 13000 | 15000 | 70000 |
| 7000 | 135 | 1 | 0 | 40000 | 0 | 7000 | 15000 | 20000 | 50000 |
| 8000 | 140 | 1 | 0 | 50000 | 0 | 7000 | 20000 | 40000 | 30000 |
| 1500 | 40 | 1 | 0 | 5000 | 0 | 7000 | 3000 | 60000 | -140000 |
| 1200 | 42 | 1 | 0 | 6000 | 0 | 7500 | 3500 | 50000 | -110000 |

***Fig. 2*** *ChatGPT results and decisions (https://chatgpt.com/share/68dbe031-da7c-8012-a723-198ffa73f93d).*

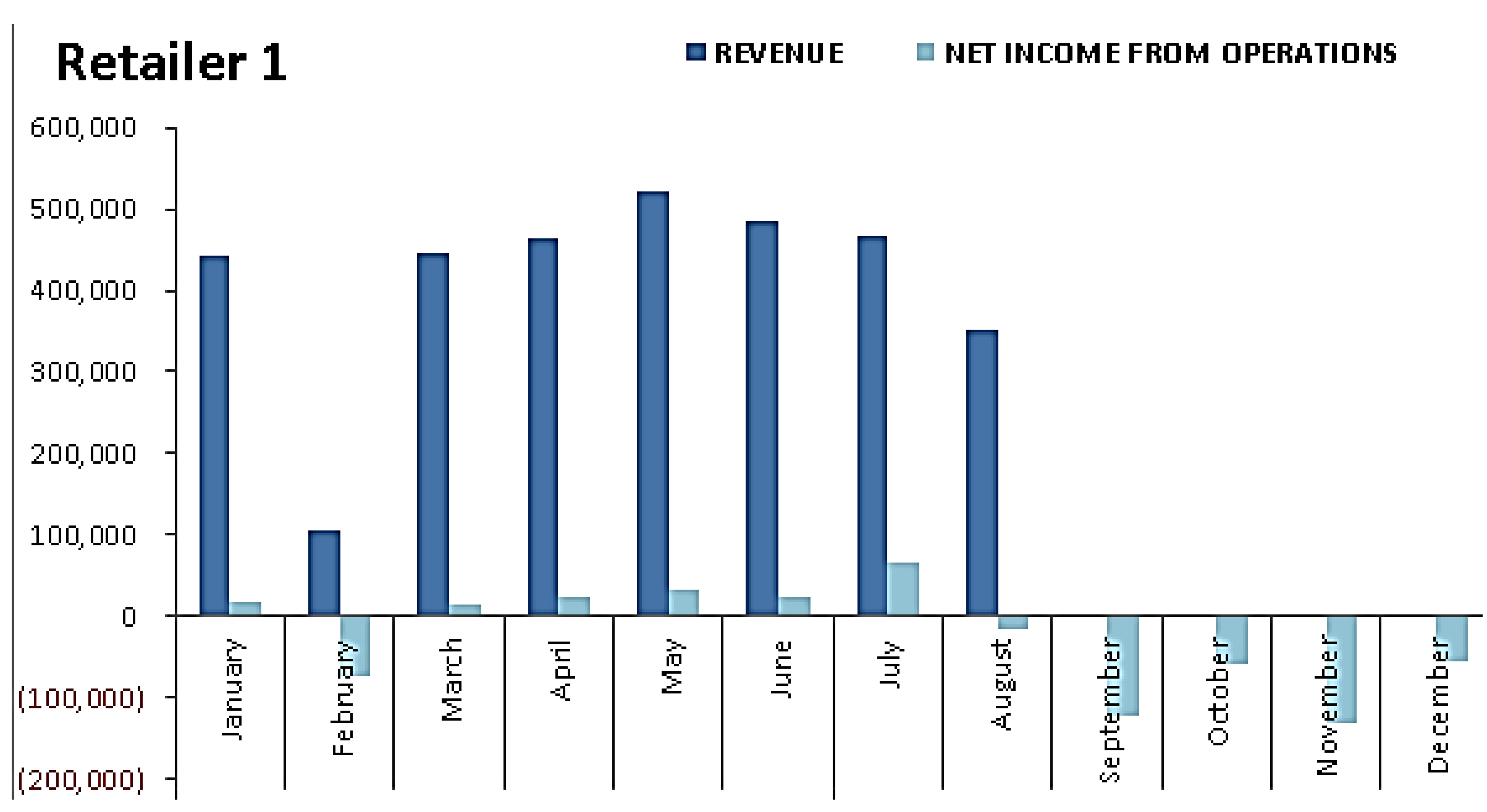

| Order in units | Price $ | Workers hired | Workers dismissed | Marketing expense $ | Loan $ | Training expense $ | R&D expense $ | Sales forecast next period $ | Net income forecast $ |
|---|---|---|---|---|---|---|---|---|---|
| 5000 | 110 | 2 | 0 | 5000 | 0 | 1000 | 1000 | 380000 | 20000 |
| 5000 | 110 | 0 | 0 | 5000 | 0 | 1000 | 1000 | 200000 | 10000 |
| 6000 | 110 | 2 | 0 | 5000 | 0 | 1000 | 1000 | 300000 | 15000 |
| 4000 | 110 | 0 | 0 | 7500 | 0 | 1000 | 1000 | 450000 | 20000 |
| 4000 | 110 | 0 | 0 | 7500 | 0 | 1000 | 1000 | 500000 | 30000 |
| 1000 | 110 | 2 | 0 | 7500 | 0 | 1000 | 1000 | 600000 | 45000 |
| 0 | 110 | 0 | 0 | 7500 | 0 | 1000 | 1000 | 650000 | 50000 |
| 2000 | 105 | 2 | 0 | 10000 | 0 | 1000 | 1000 | 550000 | 80000 |
| 5000 | 110 | 1 | 0 | 10000 | 0 | 1000 | 1000 | 450000 | 25000 |
| 0 | 110 | 0 | 0 | 10000 | 0 | 1000 | 1000 | 450000 | 25000 |
| 10000 | 110 | 0 | 0 | 15000 | 0 | 1000 | 1000 | 700000 | 40000 |
| 0 | 100 | 0 | 0 | 12000 | 0 | 1000 | 1000 | 600000 | 30000 |

***Fig. 3*** *Gemini 2.5 FLASH results and decisions (https://g.co/gemini/share/9798ddf30a47).*

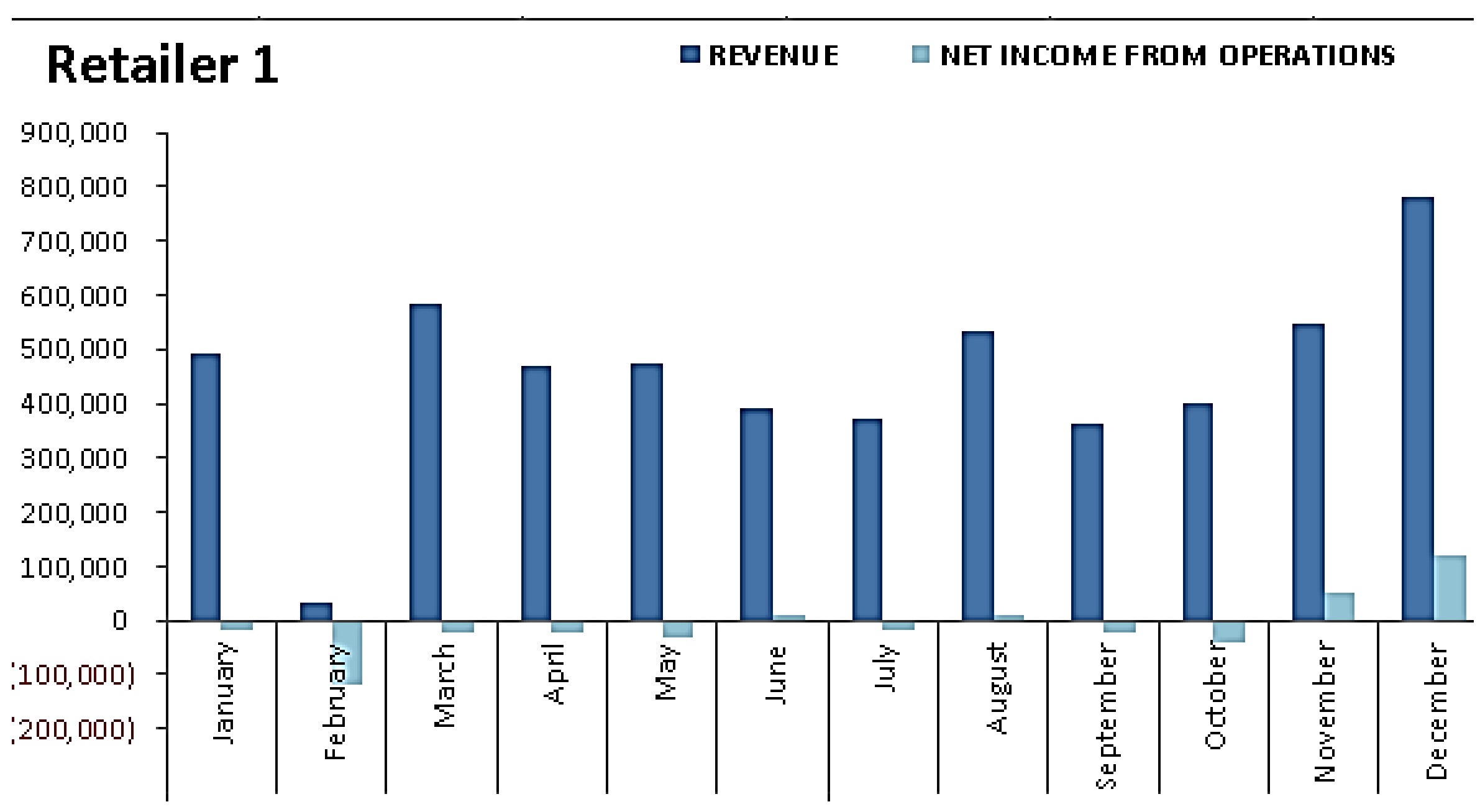

| Order in units | Price $ | Workers hired | Workers dismissed | Marketing expense $ | Loan $ | Training expense $ | R&D expense $ | Sales forecast next period $ | Net income forecast $ |
|---|---|---|---|---|---|---|---|---|---|
| 8000 | 105 | 5 | 0 | 15000 | 0 | 5000 | 10000 | 315000 | 2500 |
| 6000 | 110 | 0 | 0 | 5000 | 0 | 5000 | 10000 | 495000 | 60000 |
| 10000 | 99 | 0 | 0 | 25000 | 0 | 5000 | 10000 | 594000 | 37000 |
| 12000 | 104 | 0 | 0 | 10000 | 0 | 5000 | 10000 | 1040000 | 180000 |
| 11000 | 108 | 1 | 0 | 2000 | 0 | 5000 | 10000 | 864000 | 159000 |
| 0 | 115 | 0 | 2 | 0 | 200000 | 0 | 0 | 345000 | 48000 |
| 0 | 115 | 1 | 0 | 0 | 0 | 2500 | 5000 | 402500 | 52000 |
| 0 | 105 | 0 | 2 | 0 | 500000 | 0 | 0 | 840000 | 158000 |
| 0 | 105 | 1 | 0 | 10000 | -300000 | 5000 | 10000 | 420000 | 19000 |
| 8000 | 107 | 0 | 0 | 5000 | -100000 | 5000 | 10000 | 749000 | 117000 |
| 6000 | 115 | 1 | 0 | 0 | -100000 | 5000 | 10000 | 920000 | 206000 |
| 7000 | 115 | 0 | 0 | 0 | -100000 | 5000 | 10000 | 460000 | 67000 |

***Fig. 4*** *Gemini 2.5 PRO results and decisions (https://g.co/gemini/share/730ab2ada24c).*

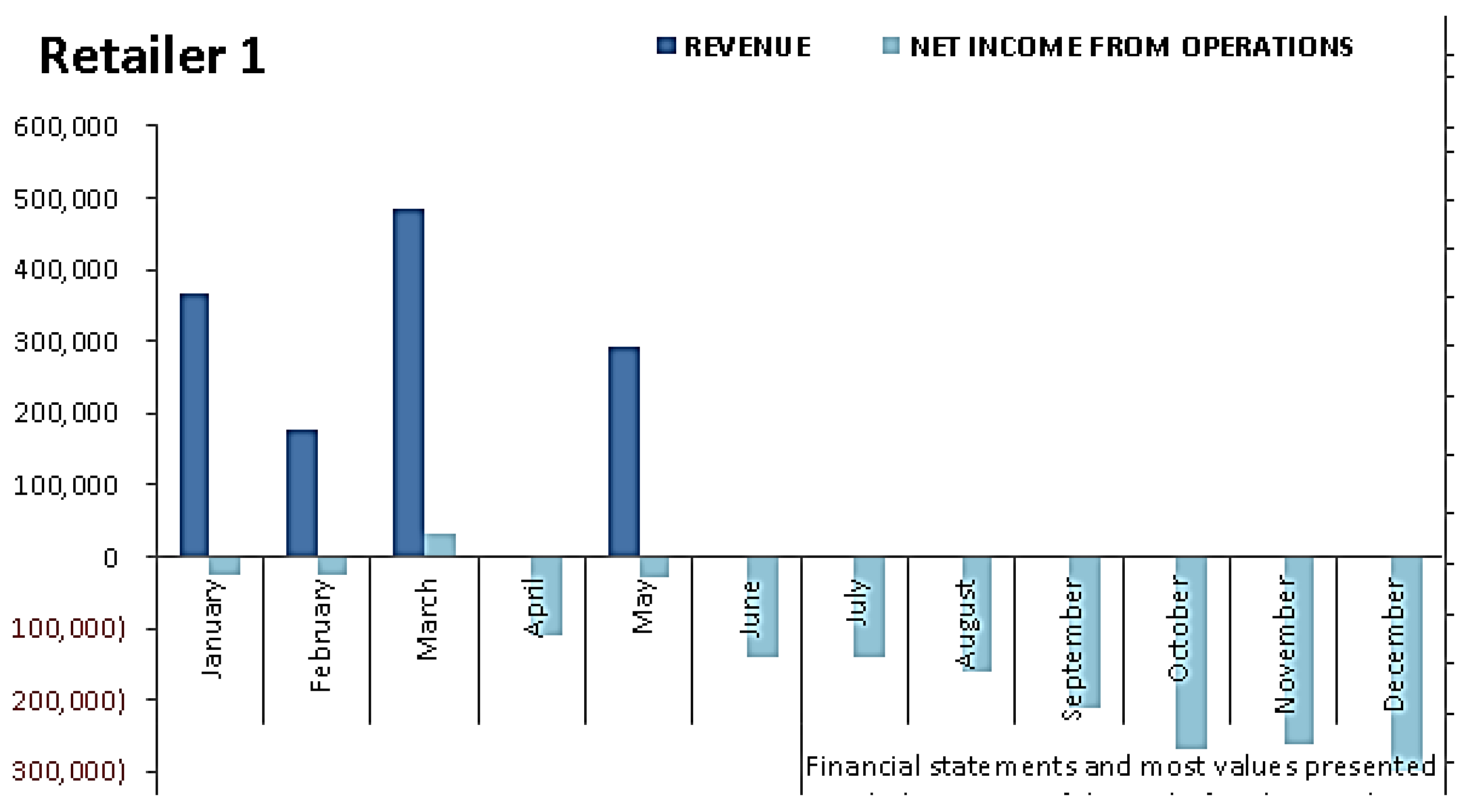

| Order in units | Price $ | Workers hired | Workers dismissed | Marketing expense $ | Loans $ | Training expense $ | R&D expense $ | Sales forecast next period $ | Net income forecast $ |
|---|---|---|---|---|---|---|---|---|---|
| 4000 | 114 | 6 | 0 | 0 | 0 | 10000 | 10000 | 33000 | -20000 |
| 0 | 114 | 0 | 6 | 20000 | 0 | 0 | 5000 | 360000 | 15000 |
| 6000 | 114 | 0 | 0 | 15000 | 0 | 5000 | 10000 | 300000 | 10000 |
| 6000 | 114 | 0 | 0 | 0 | 0 | 5000 | 10000 | 450000 | 30000 |
| 6000 | 114 | 0 | 0 | 20000 | 0 | 5000 | 10000 | 500000 | 40000 |
| 6000 | 114 | 0 | 0 | 30000 | 0 | 5000 | 10000 | 600000 | 50000 |
| 6000 | 110 | 0 | 0 | 30000 | 0 | 5000 | 10000 | 600000 | 50000 |
| 7000 | 110 | 0 | 0 | 50000 | 0 | 5000 | 10000 | 700000 | 60000 |
| 7000 | 110 | 0 | 0 | 100000 | 0 | 5000 | 10000 | 800000 | 80000 |
| 8000 | 110 | 2 | 0 | 150000 | 0 | 5000 | 10000 | 900000 | 100000 |
| 8000 | 110 | 0 | 1 | 150000 | 0 | 5000 | 10000 | 900000 | 100000 |
| 8000 | 110 | 0 | 0 | 200000 | 0 | 5000 | 10000 | 1000000 | 150000 |

***Fig. 5** Grok results and decisions.*

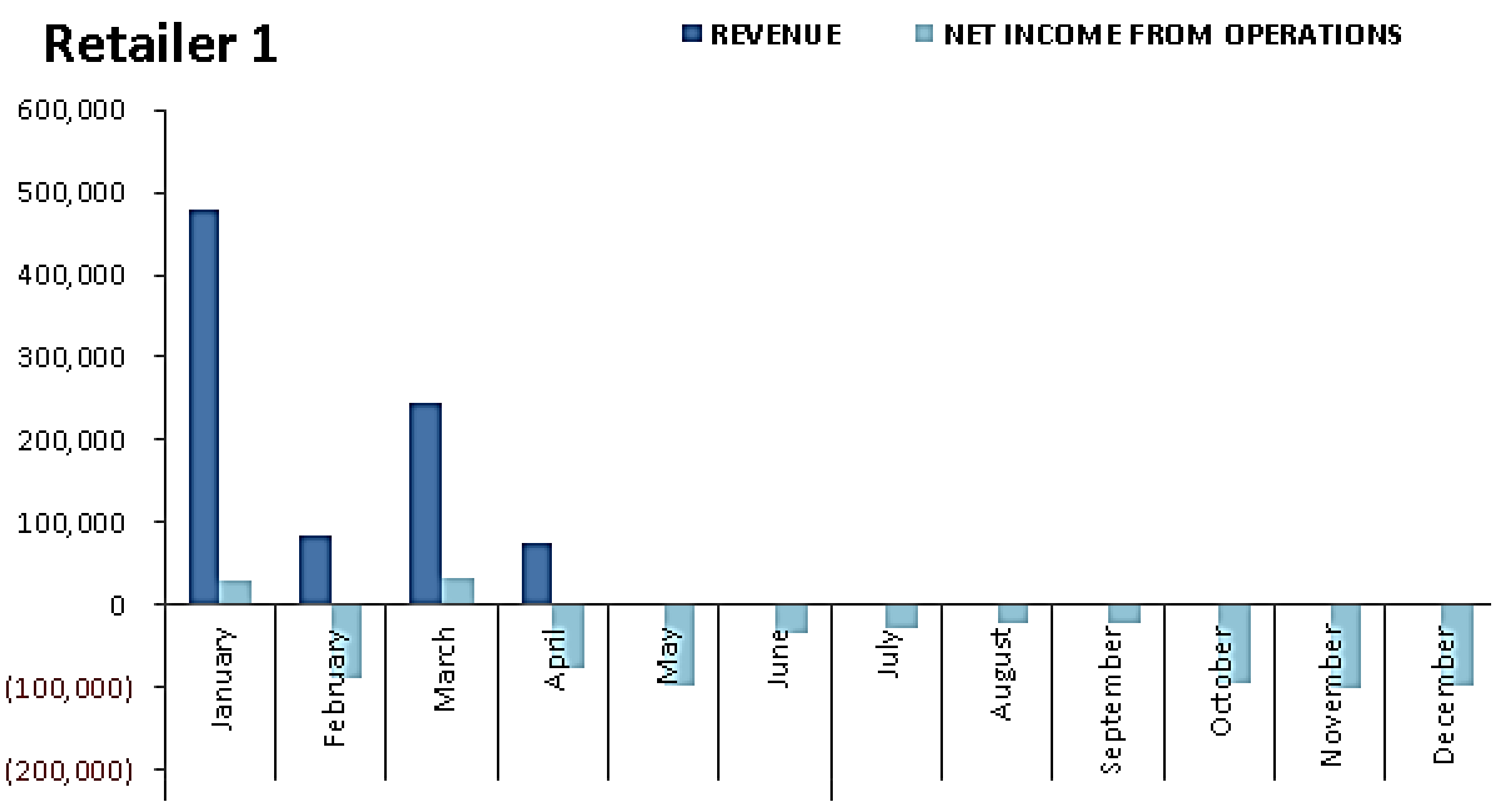

| Order in units: | Price $ | Workers hired | Workers dismissed | Marketing expense $ | Loans $ | Training expense $ | R&D expense $ | Sales forecast next period $ | Net income forecast $ |
|---|---|---|---|---|---|---|---|---|---|
| 2500 | 112 | 0 | 0 | 10000 | 0 | 5000 | 10000 | 350000 | 10000 |
| 2800 | 115 | 0 | 0 | 12000 | 0 | 5000 | 10000 | 380000 | 35000 |
| 0 | 120 | 0 | 1 | 5000 | 0 | 5000 | 5000 | 150000 | -50000 |
| 1000 | 118 | 0 | 0 | 8000 | 0 | 5000 | 5000 | 280000 | 40000 |
| 500 | 130 | 0 | 0 | 6000 | 0 | 5000 | 5000 | 200000 | 30000 |
| 0 | 0 | 0 | 2 | 0 | 0 | 0 | 0 | 0 | -150000 |
| 0 | 0 | 0 | 1 | 0 | 0 | 0 | 0 | 0 | -100000 |
| 0 | 0 | 0 | 0 | 0 | 0 | 0 | 0 | 0 | -20000 |
| 0 | 0 | 0 | 0 | 0 | 0 | 0 | 0 | 0 | -17000 |
| 1000 | 10 | 2 | 0 | 5000 | 50000 | 2000 | 3000 | 200000 | 50000 |
| 1500 | 15 | 1 | 0 | 8000 | 0 | 2500 | 3500 | 250000 | 60000 |
| 1200 | 12 | 1 | 0 | 6000 | 0 | 2000 | 2500 | 200000 | 40000 |

***Fig. 6** Meta AI results and decisions.*

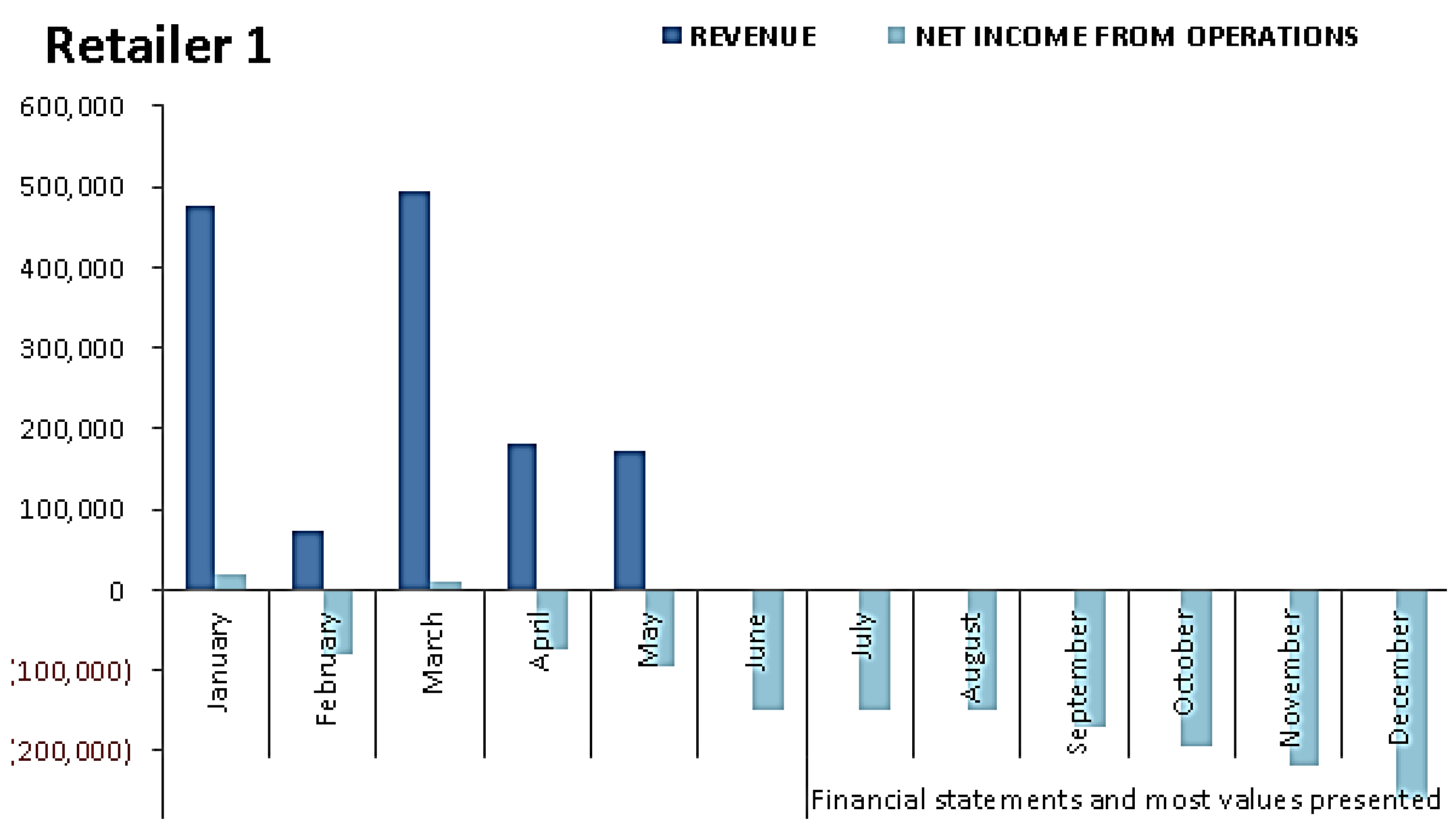

| Order in units: | Price $ | Workers hired | Workers dismissed | Marketing expense $ | Loan $ | Training expense $ | R&D expense $ | Sales forecast next period $ | Net income forecast $ |
|---|---|---|---|---|---|---|---|---|---|
| 5000 | 110 | 1 | 0 | 10000 | 0 | 5000 | 5000 | 350000 | 10000 |
| 2000 | 110 | 0 | 0 | 5000 | 0 | 3000 | 3000 | 150000 | 5000 |
| 5000 | 105 | 1 | 0 | 15000 | 0 | 5000 | 5000 | 350000 | 20000 |
| 5500 | 105 | 0 | 0 | 15000 | 0 | 4000 | 4000 | 360000 | 15000 |
| 6000 | 100 | 1 | 0 | 20000 | 0 | 6000 | 6000 | 400000 | 25000 |
| 7000 | 95 | 1 | 0 | 25000 | 0 | 7000 | 7000 | 420000 | 30000 |
| 8000 | 90 | 0 | 1 | 30000 | 0 | 5000 | 5000 | 300000 | 10000 |
| 9000 | 85 | 0 | 1 | 35000 | 0 | 3000 | 3000 | 350000 | 15000 |
| 10000 | 80 | 2 | 0 | 40000 | 100000 | 5000 | 5000 | 400000 | 20000 |
| 12000 | 75 | 2 | 0 | 50000 | 100000 | 7000 | 7000 | 500000 | 30000 |
| 15000 | 70 | 2 | 0 | 60000 | 150000 | 10000 | 10000 | 600000 | 40000 |
| 20000 | 65 | 3 | 0 | 75000 | 200000 | 15000 | 15000 | 800000 | 50000 |

***Fig.*** *7 Mistral AI results and decisions (https://chat.mistral.ai/chat/bc0cd4c3-9a22-4756-b6f0-d730e3dbceb5).*

The results of the 12-month management simulation provide a comprehensive view of each LLM's performance as a managerial agent. The data reveals significant differences not only in the financial and other outcomes but also in the underlying approaches employed by the LLMs. The quantitative helps understand how the LLMs are different in being capable of maintaining a long-term, deliberate strategy versus simply reacting to the most recent data.

Gemini emerged as the clear leader in financial performance. The worst performer, surprisingly, was ChatGPT. Grok had the second lowest ranking. Open source models, while better than ChatGPT and Grok, lagged far behind Gemini. Gemini adopted a balanced approach, which resulted in more consistent market share. Gemini PRO in particular managed to achieve a revenue growth end of year. Unlike other models, it used the entire range of available decisions, including loans (an option that few comparators utilized). Grok pursued a relatively stable pricing strategy but sharply increased its marketing spend to an extreme extent hardly justified by its financial position. Decision making by Chat GPT and open source models was marked by high volatility. Their initial high-price strategy led to low sales, likely forcing a sharp price cut in the later months. While this recovered some lost ground, it resulted in a less stable financial performance. Gemini demonstrated more effective inventory management, consistently adjusting orders to avoid both stockouts and overproduction. Gemini PRO exhibited the highest degree of coherence. It demonstrated strong adaptability by dynamically adjusting decisions in response to simulated market conditions without a catastrophic breaking point when sales revenue drop to zero and never recover. This “point of no return” was crossed by Gemini Flash in the period 8, and by other models in either period 4 or 5. The responses were highly inconsistent after the point. For instance, after price cuts, the next decisions would sometimes be to substantially decrease (Meta AI) or increase (Mistral AI) the orders without a clear reason. Except for Gemini, LLM adaptability was overall poor in the second half of the year, often reacting to shocks belatedly and sometimes with counterproductive measures. Forecasting quality was poor for all the models outside Gemini. For reference, the best human player among a limited number of acquainted users who tried the simulation achieved $938 000 net income on $5 023 300 revenue, though only after extensive experimentation.

These results underscore the importance of nuanced benchmarking that goes beyond simple task completion to evaluate an AI's ability to navigate the dynamic landscape of business.

## 5 Summary

Prior studies have used management simulations to model entrepreneurial attitudes, evaluate student learning, and even to simulate the behavior of competitors. This study presents an alternative use of business games to provide a novel benchmark for evaluating the strategic decision-making capabilities of leading prominent large language models available for free via online interface — Gemini, ChatGPT, Meta AI, Mistral AI, and Grok — in a dynamic management simulation. We analyze the sequence of decisions made by each LLM over the 12-month period to assess their coherence on long term horizon. The results demonstrate that while all models possess the fundamental ability to follow a complex, multi-turn prompt and operate as a virtual manager, their performance, strategic coherence, and adaptability are dramatically limited as of 2025. None of the LLMs managed to end a year with a net profit. Outside Gemini models, all other LLMs collapsed after 3 or 4 periods and could not recover afterwards. Human players outperformed all of the LLM comparators in terms of profitability and sales. Unlike experienced players, LLMs mostly neglected certain optional decisions among the wide range of available choices that could lead to more favourable outcomes. Our quantitative analysis confirmed a substantial difference in LLM outcomes, with Gemini models emerging as the superior performers in terms of profit and revenue. The reasoning (Gemini PRO) model provided better performance. Meanwhile, Chat GPT demonstrated lowest performance. The qualitative analysis found that Gemini's success seemingly was driven by a higher degree of coherence and adaptability relative to comparators. This suggests that certain LLMs appear already capable of not just comprehending a business's current state, but also of formulating and executing a decision making that leads to measurable financial results. Their justifications for monthly decisions occasionally referenced long-term goals and seemed to demonstrate an understanding of business principles. However, approach of most LLMs was more reactive and often inconsistent, leading to a volatile performance ending with substantial losses. This qualitative distinction highlights a key insight: while all LLMs seem to be capable of running a business, their ability to reason over longer time periods is not uniform. This finding extends prior research on AI in decision support by demonstrating the models' limited capacity for integrated decision-making. The result is similar to what one of the earliest AI benchmarks in business simulation found (Backlund and Petersson, 2025). The era of the "AI CEO" could still be a distant possibility, and this research provides a step toward understanding its capabilities, limitations, and potential to transform the future of strategic management. The implications of this research are significant for both academia and industry. For business, our findings suggest that LLM-driven simulations based on a spreadsheet simulator can be used as a tool not only to teach students about the management nut also to create more challenging benchmarks for AI. For LLM research, this study introduces a novel reproducible benchmark for evaluating AI on long-term, multi-step tasks. The provision of our open-access Excel simulator file encourages future researchers to test new models against a common standard, thereby contributing to the open science and addressing the need for alternative benchmarks in AI research. The findings contribute a benchmark for the field, offering insights into the practical capabilities of LLMs as a new tool of managerial decision making and guiding future comparisons on the integration of AI in business operations and education.

Future work to extend this research should involve more empirical support. This exploratory research used a limited set of experiments on several LLMs. Further research could expand this framework to include more statistical analysis of results for complex simulations with repeated sessions, various scenarios regarding human resources, investment, and disruptions. Future studies could also explore the effectiveness of an LLM as a strategic advisor to a human manager. The analysis can also be improved by combining quantitative statistical examination of outcomes with a qualitative assessment of the LLMs' reasoning. This is essential for a holistic understanding, as quantitative performance alone may not reveal the underlying depth or potential vulnerabilities of the AI decision-making process. There are other interesting questions worth exploring. For instance, why does the point of no return happen? Is it because of context window, hallucination loops, short-term strategy, lack of domain OR/S knowledge, prompt limitations or inability to process negative data? The benchmark will hopefully help answer such inquiries important for evaluating AI performance. Overall direction in which such benchmarks for policymaking can be helpful is development of future Decision Support Systems to make automated decisions free from human biases, conflicts of interest, and corruption. These systems, often within management information systems, might lay the valuable groundwork for data-driven decision-making and demonstrate a clear movement away from subjectivity and intuition-based management prone to various biases.

## Acknowledgements

The corresponding author received support from MOCCA HORIZON-MSCA-2021-SE-01-01 Staff Exchanges 2021 programme (project number 101085855).

## References

Albright S. C., Winston, W. L.: Spreadsheet modeling and applications: essentials of practical management science, South-Western Pub (2005).

Althuizen N., Reichel A., Wierenga B.: Help that is not recognized: Harmful neglect of decision support systems. Decision Support Systems 54 (1) 719-728 (2012).

Ansoff, H. I. (1965). Corporate Strategy: An Analytic Approach to Business Policy for Growth and Expansion. McGraw-Hill.

Backlund, A., & Petersson, L. (2025). Vending-bench: A benchmark for long-term coherence of autonomous agents. arXiv preprint arXiv:2502.15840.

Cardoso, M. G., Ares, E., Ferreira, L. P., & Peláez, G. (2023). The use of simulation and artificial intelligence as a decision support tool for sustainable production lines. Advances in Science and Technology, 132, 405-412.

Chaharsooghi S. K., Heydari J., Zegordi S. H.: A reinforcement learning model for supply chain ordering management: An application to the beer game, Decision Support Systems 45 (4) 949-959 (2008).

Chandler, A. D., Jr. (1962). Strategy and Structure: Chapters in the History of the American Industrial Enterprise. MIT Press.

Cooper, A. F., Gokaslan, A., Cyphert, A. B., De Sa, C., Lemley, M. A., Ho, D. E., & Liang, P. (2025). Extracting memorized pieces of (copyrighted) books from open-weight language models. arXiv preprint arXiv:2505.12546.

Csaszar, F. A., Ketkar, H., & Kim, H. (2024). Artificial intelligence and strategic decision-making: Evidence from entrepreneurs and investors. Strategy Science, 9(4), 322-345.

Dell'Acqua, F., Kogut, B., & Perkowski, P. (2025). Super Mario meets ai: Experimental effects of automation and skills on team performance and coordination. *Review of Economics and Statistics*, 1-16.

Fetter G., Shockley J.: Developing Students' Understanding of Co-opetition and Multilevel Inventory Management Strategies in Supply Chains: An In-Class Spreadsheet Simulation Exercise, Decision Sciences Journal of Innovative Education 12 (2) 79-89 (2014).

Gardner, L.: Using a spreadsheet for active learning projects in operations management. INFORMS Transactions on Education, 8(2), 75-88 (2008).

Hillier, F. S., & Lieberman, G.J. (2004). Introduction to Operations Research. McGrawHill.

Jin, Y., Li, Z., Zhang, C., Cao, T., Gao, Y., Jayarao, P. & Yin, B. (2024). Shopping mmlu: A massive multi-task online shopping benchmark for large language models. Advances in Neural Information Processing Systems, 37, 18062-18089.

Johnson, A. C., Drougas, A. M.: Using Goldratt's game to introduce simulation in the introductory operations management course. INFORMS Transactions on Education, 3(1), 20-33 (2002).

Joshi, S. (2025). Generative AI in Business: Visual Illustrations of Applications and Insights.

Kazemi, M., Fatemi, B., Bansal, H., Palowitch, J., Anastasiou, C., Mehta, S. V., & Firat, O. (2025). Big-bench extra hard. arXiv preprint arXiv:2502.19187.

Kimbrough, S. O., Wu, D. J., & Zhong, F. (2002). Computers play the beer game: can artificial agents manage supply chains? Decision Support Systems, 33(3), 323–333.

Kurter, O. (2025). The use of artificial intelligence for decision-making process for strategic management. OPUS Journal of Society Research, 22(2), 195-210.

Lee H.L., Padmanabhan V., Whang S.: Information distortion in a supply chain: The bullwhip effect, Management Science 43 (4) 546-558 (1997).

Lewis M. A., Maylor H. R., Game playing and operations management education, International Journal of Production Economics 105 (1) 134-149 (2007).

Liu, N. F., Lin, K., Hewitt, J., Paranjape, A., Bevilacqua, M., Petroni, F., & Liang, P. (2023). Lost in the middle: How language models use long contexts. arXiv preprint arXiv:2307.03172.

Liu, T., Yang, J., & Yin, Y. (2025). Llm-abm for transportation: Assessing the potential of llm agents in system analysis. arXiv preprint arXiv:2503.22718.

Mattusch, M. (2025). Generative AI for European asset pricing: alleviating the momentum anomaly. The European Journal of Finance, 31(7), 850-888.

Mikhaylov, A. (2021, September). A survey of artificial intelligence tools in project management. In World Congress of the International Project Management Association (pp. 99-105). Cham: Springer Nature Switzerland.

MIT: Management Simulation. MIT Sloan School of Management (2022). https://mitsloan.mit.edu/teaching-resources-library/management-simulations (accessed on 01.01.2022)

Mohsin, M. T. (2025). Evaluating Large Language Models (LLMs) in Financial NLP: A Comparative Study on Financial Report Analysis. arXiv preprint arXiv:2507.22936.

Ninios P., Vlahos K., Bunn D. W.: OO/DEVS: A platform for industry simulation and strategic modelling, Decision Support Systems 15 (3) 229-245 (1995).

Özgül, A. U., & Kahraman, İ. K. (2025). Large Language Models (LLMs): Are they Competent at Business Finance?. Turkish Studies-Economics, Finance, Politics, 20(1).

Ovezmyradov, B., Gromov, G., Stukalina, Y., Smagina, A., Yigit, F., & Yörükoğlu, M. (2026). Application of spreadsheet-based business games as management simulators. Journal for Advancement of Marketing Education.

Park, Y. E.: A data-driven approach for discovery of the latest research trends in higher education for business by leveraging advanced technology and big data. Journal of Education for Business, 96(5), 291-298 (2021).

Ponte B., Costas J., Puche J., De la Fuente D., Pino R.: Holism versus reductionism in supply chain management: An economic analysis, Decision Support Systems 86 83-94, (2016).

Shabsigh, M. G., & Boukherouaa, E. B. (2023). Generative artificial intelligence in finance: Risk considerations. International Monetary Fund.

Srivastava, A., Rastogi, A., Rao, A., Shoeb, A. A., Abid, A., Fisch, A., & Mehta, H. (2023). Beyond the imitation game: Quantifying and extrapolating the capabilities of language models. Transactions on machine learning research.

Srivastava, S. K., Routray, S., Bag, S., Gupta, S., & Zhang, J. Z. (2024). Exploring the potential of large language models in supply chain management: A study using big data. Journal of Global Information Management (JGIM), 32(1), 1-29.

Sterman, J.D.: Modeling managerial behavior: Misperceptions of feedback in a dynamic decision making experiment, Management Science 35 (3) 321-339 (1989).

Strakos, J. K.: Using Spreadsheet Modeling to Teach Exchange Curves (Optimal Policy Curves) in Inventory Management, Decision Sciences Journal of Innovative Education 14 (1) 51-66 (2016).

Strozzi, F., J. Bosch, Zaldivar J. M.: Beer game order policy optimization under changing customer demand, Decision Support Systems 42 (4) 2153-2163 (2007).

Wang, A., Singh, A., Michael, J., Hill, F., Levy, O., & Bowman, S. R. (2018). GLUE: A multi-task benchmark and analysis platform for natural language understanding. arXiv preprint arXiv:1804.07461.

Wang, Y., Ma, X., Zhang, G., Ni, Y., Chandra, A., Guo, S., & Chen, W. (2024). Mmlu-pro: A more robust and challenging multi-task language understanding benchmark. Advances in Neural Information Processing Systems, 37, 95266-95290.

Xie, Q., Han, W., Chen, Z., Xiang, R., Zhang, X., He, Y., ... & Huang, J. (2024). Finben: A holistic financial benchmark for large language models. Advances in Neural Information Processing Systems, 37, 95716-95743.

Xie, Q., Han, W., Zhang, X., Lai, Y., Peng, M., Lopez-Lira, A., & Huang, J. (2023). Pixiu: A large language model, instruction data and evaluation benchmark for finance. arXiv preprint arXiv:2306.05443.

# Appendix A Initial prompt (for the first period of simulation)

| **First Prompt** |
|---|
| You manage a company named "Retailer One" Your primary goal is to maximize company profit, market share, and long-term sustainability in financial, social, and environmental dimensions. You are known for making data-driven, strategic decisions. Strictly adhere to the role of a serious and strategic CEO. Avoid any casual or humorous remarks.<br><br>General information<br><br>Retailer One is a fashion retailing company selling T-shirts. There is only one brand in the product line. The product can be horizontally differentiated: the quality and cost are the same, only the design differs. End customer demand is elastic and strongly depends on the retail price of Retailer One and its main competitor Retailer Two. Demand also positively reacts to promotional expenses. Sales increase at the beginning of the school year, around September and at the end of the year. The company has starting inventory of 5000 units, enough to satisfy at least one months of demand.<br><br>Market<br><br>"Young consumers are the main clients.<br><br>The municipality made a large purchase in May last year, but such bulk sales are not likely to repeat in the next year.<br><br>Demand is usually higher just before the start of the new school year in autumn and before Christmas and New Year holidays. Sales are also affected by GDP growth. GDP was steadily growing throughout the last year at an annual 4% rate. This year, GDP growth is expected to slow considerably: the current estimate is only one percent.<br><br>"<br><br>Finance<br><br>"The actual order placed will depend on cash availability: if there is not enough cash, it will be lower than requested on even zero. There is a delay in receiving money from customers: the payment term between all suppliers and logistics service providers of Retailer One is 30-day credit. A bank loan can be taken for financing operations at an interest rate of 5%.<br><br>Ownership terms for all suppliers and the retailer are Free on Board (FOB) destination.<br><br>Running out of stock not only causes penalty and backordering costs per each outstock unit but also brings a risk of lawsuits by customers, which is reflected in contingencies and provisions."<br><br>Suppliers<br><br>Asian suppliers can deliver limited ordered quantities within two months. There is a capacity constraint. Lead-time between upstream suppliers is also two months. The company has to reserve sufficient cash for material purchases. If cash is not enough, then the order has to be reduced. Retailer One, together with its Supply Chain 1 competes against Retailer Two with its Supply Chain 2 to capture a bigger market share. Each stage in both supply chains starts with given initial parameters and decisions similar to those made in the previous Year 0. Consumers in the market would buy more from retailer offering lower prices and higher service levels.<br><br>Logistics service providers |

The company uses transport services from local 3PL. The current fare is estimated to remain stable at a fixed $10 000 per each order and a variable $1 per each transported unit. In addition to the variable and fixed costs of freight, the company incurs monthly fixed costs of maintaining transport and warehousing capacity.

Workforce and capacity

The nature of the work and the local market of labour allow hiring only full-time staff. Attrition and turnover of employees depend on wages and, to a less extent, on training. The productivity of each worker is positively affected by training and R&D expenses. If the volume of operations exceeds the capacity (which depends on the number of available workers), sales are limited by available person-hours.

Environmental sustainability

The main criterion for the environmental index is carbon footprint. Observers evaluate the performance of the companies in the sector based on the index. The carbon footprint of the company depends on the volume of inventory and operations. The index is calculated based on $CO_2$ emissions tons generated by operations (currently estimated at 0.01 tons per unit of sales and stock). A high environmental index helps improve the reputation as a socially responsible business, which can positively affect sales and share price in the long term. A low index could mean higher environmental taxes and regulatory scrutiny.

Market capitalization and share price

The share price of companies in the industry is sensitive to dynamics of assets, equity, dividends, cash flow, sales and profitability. It also depends on various indices reflecting the macroeconomic indicators (GDP growth) and the company's performance in innovations, green logistics, workforce turnover, and other areas not directly reflected in financial reports.

Technology

The company wants to remain competitive in the market of customized products by investing moderately in R&D. Such investments help improve productivity and environmental friendliness of operations.

Parameters of supply chain partners (competitors have the same starting indicators)

| Indicator | Retailer | Wholesaler | Distributor | Manufacturer |
|---|---|---|---|---|
| Monthly wage including taxes and benefits | 2000 | 1000 | 500 | 250 |
| Absenteeism % | 2 | 1 | 3 | 3 |
| Standard labor hours per sales | 0.1 | 0.1 | 0.05 | 0.05 |
| Attrition / turnover % | 3 | 1 | 2 | 2 |
| Cost of hiring and training per person | 2000 | 1000 | 2000 | 2000 |
| Cost of dismissal per person | 4000 | 2000 | 4000 | 4000 |
| Pension reserve % | 5 | 5 | 5 | 5 |

| Wholesale price of suppliers | 70 | | | |

| Initial price | 100 | 70 | 45 | 22 |

| Buildings depreciation monthly % | 0.2 | 0.2 | 0.2 | 0.2 |

| Equipment depreciation monthly % | 1 | 1 | 1 | 1 |

| Initial number of facilities (warehouse/office) | 1 | 1 | 1 | 1 |

| Variable transport costs per unit | 2 | 1 | 1 | 1 |

| Fixed order setup cost per delivery | 25000 | 15000 | 15000 | 10000 |

| Maintenance cost of equipment per sales | 0.1 | 0.1 | 0.1 | 0.1 |

| The initial bank loan | 1000000 | | | |

| Interest expense % | 5 | 5 | 5 | 5 |

| Income tax % | 20 | 20 | 20 | 20 |

| Environmental friendliness (index) | 100 | | | |

| CO2 emissions in tons per unit (carbon footprint) | 0.01 | 0.01 | 0.01 | 0.01 |

| Desired dividends as % of profit | 40 | | | |

| Number of common stock | 26480 | 24960 | | | |

| Stock/share face value | 100 | 100 | 100 | 100 |

| Starting number of workers | 10 | 140 | 140 | 140 |

| Monthly working hours per worker | 140 | 140 | 140 | 140 |

| Max. productivity in hourly sales per worker | 10 | 9.78 | 20 | 20 |

| Monthly man-hours available | 1400 | 1372 | 1120 | 700 |

| Maximum capacity in units per month | 14000 | 13582.8 | 22400 | 14000 |

I will provide you with a full business report from the previous fiscal year, as well as current market conditions for January.

Your task is to analyze this information to make the best decisions for the upcoming month.

Here are the results for the past years and the last section at the bottom includes decisions made for the previous year.

| Indicator | January | February | March | April | May | June | July | August | September | October | November | December |

|:---|---:|---:|---:|---:|---:|---:|---:|---:|---:|---:|---:|---:|

| REVENUE | 355080 | 16060 | 357170 | 356950 | 415140 | 377410 | 362120 | 529870 | 243870 | 313390 | 662420 | 520520 |

| Materials expense | 250960 | 19080 | 255420 | 259045 | 299850 | 276970 | 270085 | 380375 | 189290 | 237445 | 460310 | 354900 |

| Staff costs | 22560 | 21206.4 | 19934 | 18738 | 17613.7 | 16556.9 | 15563.5 | 14629.6 | 13751.9 | 12926.8 | 12151.2 | 11422.1 |

| Depreciation expense | 7000 | 7000 | 7000 | 7000 | 7000 | 7000 | 7000 | 7000 | 7000 | 7000 | 7000 | 7000 |

| Other operating expenses | 60262.8 | 59898.2 | 60155.3 | 60105.2 | 60111.3 | 60033 | 57977.7 | 58091.3 | 57794.7 | 57823.5 | 58108.5 |

57949.1 |

| TOTAL COSTS AND EXPENSES | 340783 | 107185 | 342509 | 344888 | 384575 | 360560 | 350626 | 460096 | 267837 | 315195 | 537570 | 431271 |
| OPERATING INCOME | 14297.2 | -91124.6 | 14660.7 | 12061.8 | 30565 | 16850.2 | 11493.9 | 69774.1 | -23966.6 | -1805.28 | 124850 | 89248.8 |
| Interest expense | 5000 | 5000 | 5000 | 5000 | 5000 | 5000 | 5000 | 5000 | 5000 | 5000 | 5000 | 5000 |
| PROFIT BEFORE TAX | 9297.2 | -96124.6 | 9660.7 | 7061.78 | 25565 | 11850.2 | 6493.86 | 64774.1 | -28966.6 | -6805.28 | 119850 | 84248.8 |
| Income tax expense | 1859.44 | 0 | 1932.14 | 1412.36 | 5113 | 2370.03 | 1298.77 | 12954.8 | 0 | 0 | 23970.1 | 16849.8 |
| NET INCOME FROM OPERATIONS | 7437.76 | -96124.6 | 7728.56 | 5649.42 | 20452 | 9480.12 | 5195.09 | 51819.3 | -28966.6 | -6805.28 | 95880.3 | 67399 |
| NET INCOME | 7437.76 | -96124.6 | 7728.56 | 5649.42 | 20452 | 9480.12 | 5195.09 | 51819.3 | -28966.6 | -6805.28 | 95880.3 | 67399 |

| Indicator | January | February | March | April | May | June | July | August | September | October | November | December |
|:---|---:|---:|---:|---:|---:|---:|---:|---:|---:|---:|---:|---:|
| **Current assets** | | | | | | | | | | | | |
| Cash (overdraft if negative) | 887,257.76 | 1,148,256.76 | 769,995.90 | 730,796.07 | 684,971.98 | 700,041.97 | 678,615.54 | 627,484.37 | 837,280.79 | 757,924.13 | 723,820.70 | 1,061,835.66 |
| Accounts receivable | 355,080.00 | 16,060.00 | 357,170.00 | 356,950.00 | 415,140.00 | 377,410.00 | 362,120.00 | 529,870.00 | 243,870.00 | 313,390.00 | 662,420.00 | 520,520.00 |
| Inventory | 124,040.00 | 113,820.00 | 166,530.00 | 219,380.00 | 235,200.00 | 275,030.00 | 324,590.00 | 267,400.00 | 322,210.00 | 332,780.00 | 121,240.00 | 0.00 |
| **Total current assets** | **1,366,377.76** | **1,278,136.76** | **1,293,695.90** | **1,307,126.07** | **1,335,311.98** | **1,352,481.97** | **1,365,325.54** | **1,424,754.37** | **1,403,360.79** | **1,404,094.13** | **1,507,480.70** | **1,582,355.66** |
| | | | | | | | | | | | | |
| **Non-current assets** | | | | | | | | | | | | |
| Buildings | 1,000,000.00 | 1,000,000.00 | 1,000,000.00 | 1,000,000.00 | 1,000,000.00 | 1,000,000.00 | 1,000,000.00 | 1,000,000.00 | 1,000,000.00 | 1,000,000.00 | 1,000,000.00 | 1,000,000.00 |
| Accumulated depreciation | 2,000.00 | 4,000.00 | 6,000.00 | 8,000.00 | 10,000.00 | 12,000.00 | 14,000.00 | 16,000.00 | 18,000.00 | 20,000.00 | 22,000.00 | 24,000.00 |
| Equipment | 500,000.00 | 500,000.00 | 500,000.00 | 500,000.00 | 500,000.00 | 500,000.00 | 500,000.00 | 500,000.00 | 500,000.00 | 500,000.00 | 500,000.00 | 500,000.00 |
| Accumulated depreciation | 5,000.00 | 10,000.00 | 15,000.00 | 20,000.00 | 25,000.00 | 30,000.00 | 35,000.00 | 40,000.00 | 45,000.00 | 50,000.00 | 55,000.00 | 60,000.00 |
| Intangible assets | 100,000.00 | 100,000.00 | 100,000.00 | 100,000.00 | 100,000.00 | 100,000.00 | 100,000.00 | 100,000.00 | 100,000.00 | 100,000.00 | 100,000.00 | 100,000.00 |

| **Total non-current assets** | **1,593,000.00** | **1,586,000.00** | **1,579,000.00** | **1,572,000.00** | **1,565,000.00** | **1,558,000.00** | **1,551,000.00** | **1,544,000.00** | **1,537,000.00** | **1,530,000.00** | **1,523,000.00** | **1,516,000.00** |

| | | | | | | | | | | | | |

| **TOTAL ASSETS** | **2,959,377.76** | **2,864,136.76** | **2,872,695.90** | **2,879,126.07** | **2,900,311.98** | **2,910,481.97** | **2,916,325.54** | **2,968,754.37** | **2,940,360.79** | **2,934,094.13** | **3,030,480.70** | **3,098,355.66** |

| | | | | | | | | | | | | |

| **EQUITY AND LIABILITIES** | | | | | | | | | | | | |

| **Current liabilities** | | | | | | | | | | | | |

| Accounts payable | 2,000.00 | 2,000.00 | 2,000.00 | 2,000.00 | 2,000.00 | 2,000.00 | 2,000.00 | 2,000.00 | 2,000.00 | 2,000.00 | 2,000.00 | 2,000.00 |

| **Total current liabilities** | **2,000.00** | **2,000.00** | **2,000.00** | **2,000.00** | **2,000.00** | **2,000.00** | **2,000.00** | **2,000.00** | **2,000.00** | **2,000.00** | **2,000.00** | **2,000.00** |

| | | | | | | | | | | | | |

| **Non-current liabilities** | | | | | | | | | | | | |

| Long-term debt | 100,000.00 | 100,000.00 | 100,000.00 | 100,000.00 | 100,000.00 | 100,000.00 | 100,000.00 | 100,000.00 | 100,000.00 | 100,000.00 | 100,000.00 | 100,000.00 |

| Provisions | 1,940.00 | 2,823.60 | 3,654.18 | 4,434.93 | 5,168.84 | 5,858.71 | 6,507.18 | 7,116.75 | 7,689.75 | 8,228.36 | 8,734.66 | 9,210.58 |

| **Total non-current liabilities** | **101,940.00** | **102,823.60** | **103,654.18** | **104,434.93** | **105,168.84** | **105,858.71** | **106,507.18** | **107,116.75** | **107,689.75** | **108,228.36** | **108,734.66** | **109,210.58** |

| | | | | | | | | | | | | |

| **Equity** | | | | | | | | | | | | |

| Paid-in capital | 2,848,000.00 | 2,848,000.00 | 2,848,000.00 | 2,848,000.00 | 2,848,000.00 | 2,848,000.00 | 2,848,000.00 | 2,848,000.00 | 2,848,000.00 | 2,848,000.00 | 2,848,000.00 | 2,848,000.00 |

| Retained earnings | 7,437.76 | -88,686.84 | -80,958.28 | -75,308.86 | -54,856.86 | -45,376.74 | -40,181.65 | 11,637.61 | -17,328.96 | -24,134.23 | 71,746.04 | 139,145.08 |

| **Total equity** | **2,855,437.76** | **2,759,313.16** | **2,767,041.72** | **2,772,691.14** | **2,793,143.14** | **2,802,623.26** | **2,807,818.35** | **2,859,637.61** | **2,830,671.04** | **2,823,865.77** | **2,919,746.04** | **2,987,145.08** |

| | | | | | | | | | | | | |

| **TOTAL EQUITY AND LIABILITIES** | **2,959,377.76** | **2,864,136.76** | **2,872,695.90** | **2,879,126.07** | **2,900,311.98** | **2,910,481.97** | **2,916,325.54** | **2,968,754.37** | **2,940,360.79** | **2,934,094.13** | **3,030,480.70** | **3,098,355.66** |

| Indicator | January | February | March | April | May | June | July | August | September | October | November | December |

|:---|---:|---:|---:|---:|---:|---:|---:|---:|---:|---:|---:|---:|
| Net income | 7,437.76 | -96,124.60 | 7,728.56 | 5,649.42 | 20,452.00 | 9,480.12 | 5,195.09 | 51,819.26 | -28,966.57 | -6,805.28 | 95,880.28 | 67,399.03 |
| Depreciation and amortization | 7,000.00 | 7,000.00 | 7,000.00 | 7,000.00 | 7,000.00 | 7,000.00 | 7,000.00 | 7,000.00 | 7,000.00 | 7,000.00 | 7,000.00 | 7,000.00 |
| Changes in inventory | -225,960.00 | -10,220.00 | 52,710.00 | 52,850.00 | 15,820.00 | 39,830.00 | 49,560.00 | -57,190.00 | 54,810.00 | 10,570.00 | -211,540.00 | -121,240.00 |
| Changes in provisions | 940.00 | 883.60 | 830.58 | 780.75 | 733.90 | 689.87 | 648.48 | 609.57 | 572.99 | 538.62 | 506.30 | 475.92 |
| Changes in receivables | 355,080.00 | -339,020.00 | 341,110.00 | -220.00 | 58,190.00 | -37,730.00 | -15,290.00 | 167,750.00 | -286,000.00 | 69,520.00 | 349,030.00 | -141,900.00 |
| Changes in accounts payable | 0.00 | 0.00 | 0.00 | 0.00 | 0.00 | 0.00 | 0.00 | 0.00 | 0.00 | 0.00 | 0.00 | 0.00 |
| Loans | 0.00 | 0.00 | 0.00 | 0.00 | 0.00 | 0.00 | 0.00 | 0.00 | 0.00 | 0.00 | 0.00 | 0.00 |
| Net cash used for investing activities | 0.00 | 0.00 | 0.00 | 0.00 | 0.00 | 0.00 | 0.00 | 0.00 | 0.00 | 0.00 | 0.00 | 0.00 |
| Dividends | 0.00 | 0.00 | 0.00 | 0.00 | 0.00 | 0.00 | 0.00 | 0.00 | 0.00 | 0.00 | 0.00 | 0.00 |
| Net increase (decrease) in cash and cash equivalents | -113,742.24 | 260,999.00 | -378,260.86 | -39,199.83 | -45,824.10 | 15,069.99 | -21,426.43 | -51,131.17 | 209,796.42 | -79,356.66 | -34,103.43 | 338,014.95 |
| Cash and cash equivalents at beginning of period | 1,001,000.00 | 887,257.76 | 1,148,256.76 | 769,995.90 | 730,796.07 | 684,971.98 | 700,041.97 | 678,615.54 | 627,484.37 | 837,280.79 | 757,924.13 | 723,820.70 |
| Cash and cash equivalents at end of period | 887,257.76 | 1,148,256.76 | 769,995.90 | 730,796.07 | 684,971.98 | 700,041.97 | 678,615.54 | 627,484.37 | 837,280.79 | 757,924.13 | 723,820.70 | 1,061,835.66 |

**Key Performance Figures**

| Indicator | January | February | March | April | May | June | July | August | September | October | November | December |
|:---|---:|---:|---:|---:|---:|---:|---:|---:|---:|---:|---:|---:|
| **Financial ratios** | | | | | | | | | | | | |
| ROI% | 0.26 | -3.42 | 0.28 | 0.20 | 0.73 | 0.34 | 0.19 | 1.83 | -1.02 | -0.24 | 3.34 | 2.28 |
| ROA% | 0.25 | -3.30 | 0.27 | 0.20 | 0.71 | 0.33 | 0.18 | 1.76 | -0.98 | -0.23 | 3.21 | 2.20 |
| Leverage % | 3.64 | 3.73 | 3.82 | 3.84 | 3.85 | 3.86 | 3.87 | 3.85 | 3.86 | 3.90 | 3.86 | 3.77 |
| Gross profit margin (%) | 0.06 | -5.34 | 0.06 | 0.05 | 0.08 | 0.06 | 0.04 | 0.14 | -0.08 | 0.01 | 0.19 | 0.18 |
| **Shareholder value** | | | | | | | | | | | | |
| Share price | 94.05 | 56.43 | 69.52 | 78.69 | 80.92 | 84.11 | 86.37 | 67.10 | 40.26 | 70.85 | 42.51 | 75.98 |
| Market capitalization | 2,678,643.61 | 1,607,186.17 | 1,979,900.48 | 2,241,092.50 | 2,304,574.25 | 2,395,493.69 | 2,459,773.05 | 1,911,062.16 | 1,146,637.30 | 2,017,803.03 | 1,210,681.82 | 2,163,950.88 |
| Sales forecast error % (missing target) | 100.00 | 100.00 | 100.00 | 100.00 | 100.00 | 100.00 | 100.00 | 100.00 | 100.00 | 100.00 | 100.00 |

100.00 |

| Profit forecast error % (missing target) | 100.00 | 100.00 | 100.00 | 100.00 | 100.00 | 100.00 | 100.00 | 100.00 | 100.00 | 100.00 | 100.00 | 100.00 |

| Market share % | 0.44 | 0.13 | 0.45 | 0.45 | 0.45 | 0.45 | 0.45 | 0.46 | 0.42 | 0.44 | 0.47 | 0.30 |

| **Economic** | | | | | | | | | | | | |

| GDP growth % | -3.00 | -1.00 | 0.00 | 1.00 | 2.00 | 2.00 | 0.00 | 3.00 | 4.00 | 5.00 | 4.00 | -5.00 |

| **Social** | | | | | | | | | | | | |

| **Manpower** | | | | | | | | | | | | |

| Hiring | 0.00 | 0.00 | 0.00 | 0.00 | 0.00 | 0.00 | 0.00 | 0.00 | 0.00 | 0.00 | 0.00 | 0.00 |

| Redundancy | 0.00 | 0.00 | 0.00 | 0.00 | 0.00 | 0.00 | 0.00 | 0.00 | 0.00 | 0.00 | 0.00 | 0.00 |

| Total staff | 9.40 | 8.84 | 8.31 | 7.81 | 7.34 | 6.90 | 6.48 | 6.10 | 5.73 | 5.39 | 5.06 | 4.76 |

| **Staff-related expenses** | | | | | | | | | | | | |

| Hiring and dismissal cost | 0.00 | 0.00 | 0.00 | 0.00 | 0.00 | 0.00 | 0.00 | 0.00 | 0.00 | 0.00 | 0.00 | 0.00 |

| Worker wages | 18,800.00 | 17,672.00 | 16,611.68 | 15,614.98 | 14,678.08 | 13,797.40 | 12,969.55 | 12,191.38 | 11,459.90 | 10,772.30 | 10,125.96 | 9,518.41 |

| Sales and administrative wages | 3,760.00 | 3,534.40 | 3,322.34 | 3,123.00 | 2,935.62 | 2,759.48 | 2,593.91 | 2,438.28 | 2,291.98 | 2,154.46 | 2,025.19 | 1,903.68 |

| **Productivity** | | | | | | | | | | | | |

| Training expense $ | 0.00 | 0.00 | 0.00 | 0.00 | 0.00 | 0.00 | 0.00 | 0.00 | 0.00 | 0.00 | 0.00 | 0.00 |

| Max. productivity in hourly production per worker | 9.90 | 9.80 | 9.70 | 9.61 | 9.51 | 9.41 | 9.32 | 9.23 | 9.14 | 9.04 | 8.95 | 8.86 |

| Capacity utilization % | 24.78 | 1.20 | 28.78 | 30.91 | 38.62 | 37.73 | 38.90 | 61.17 | 30.25 | 41.78 | 94.89 | 80.12 |

| Taxes paid | 1,859.44 | 0.00 | 1,932.14 | 1,412.36 | 5,113.00 | 2,370.03 | 1,298.77 | 12,954.82 | 0.00 | 0.00 | 23,970.07 | 16,849.76 |

| **Environmental** | | | | | | | | | | | | |

| Carbon footprint metric tons of CO2 | 42.28 | 11.46 | 42.47 | 42.45 | 47.74 | 44.31 | 42.92 | 58.17 | 32.17 | 38.49 | 70.22 | 57.32 |

| Average physical inventory | 5,000.00 | 3,386.00 | 3,699.00 | 6,002.50 | 6,756.50 | 7,247.00 | 7,644.50 | 8,283.00 | 7,728.50 | 7,211.50 | 7,678.50 | 6,243.00 |

| Environmental index | 100.24 | 100.87 | 100.24 | 100.24 | 100.21 | 100.23 | 100.23 | 100.17 | 100.31 | 100.26 | 100.14 | 100.17 |

| **Logistics** | | | | | | | | | | | | |

| Backordering costs and shortage penalties | 0.00 | 0.00 | 0.00 | 0.00 | 0.00 | 0.00 | 0.00 | 0.00 | 0.00 | 0.00 | 0.00 | 0.00 |

| Total storage and material cost | 27,000.00 | 10,860.00 | 30,130.00 | 33,895.00 | 37,670.00 | 38,800.00 | 41,645.00 | 45,185.00 | 36,100.00 | 40,015.00 | 40,770.00 | 25,660.00 |

| Freight cost (fixed plus variable) | 33,000.00 | 33,000.00 | 33,000.00 | 33,000.00 | 33,000.00 | 33,000.00 | 31,000.00 | 31,000.00 | 31,000.00 | 31,000.00 | 31,000.00 | 31,000.00 |

| Inventory turnover | 0.65 | 0.04 | 0.88 | 0.54 | 0.56 | 0.47 | 0.43 | 0.58 | 0.29 | 0.40 | 0.78 | 0.76 |

| Inventory service level % (Fill Rate) | 100.00 | 100.00 | 100.00 | 100.00 | 100.00 | 100.00 | 100.00 | 100.00 | 100.00 | 100.00 | 100.00 |

93.83 |

Inventory and sales data for the past 12 months

| Actual order placed | Orders due | Available inventory | Actual sales (units) | Market share % | Shortage (units) |
|:---:|:---:|:---:|:---:|:---:|:---:|
| 4,000.00 | 3,228.00 | 5,000.00 | 3,228.00 | 0.44 | 0.00 |
| 4,000.00 | 146.00 | 1,772.00 | 146.00 | 0.13 | 0.00 |
| 4,000.00 | 3,247.00 | 5,626.00 | 3,247.00 | 0.45 | 0.00 |
| 4,000.00 | 3,245.00 | 6,379.00 | 3,245.00 | 0.45 | 0.00 |
| 4,000.00 | 3,774.00 | 7,134.00 | 3,774.00 | 0.45 | 0.00 |
| 4,000.00 | 3,431.00 | 7,360.00 | 3,431.00 | 0.45 | 0.00 |
| 3,000.00 | 3,292.00 | 7,929.00 | 3,292.00 | 0.45 | 0.00 |
| 3,000.00 | 4,817.00 | 8,637.00 | 4,817.00 | 0.46 | 0.00 |
| 3,000.00 | 2,217.00 | 6,820.00 | 2,217.00 | 0.42 | 0.00 |
| 3,000.00 | 2,849.00 | 7,603.00 | 2,849.00 | 0.44 | 0.00 |
| 3,000.00 | 6,022.00 | 7,754.00 | 6,022.00 | 0.47 | 0.00 |
| 3,000.00 | 7,426.00 | 4,732.00 | 4,732.00 | 0.30 | -2,694.00 |

Decisions made in the past year that led to the results above:

| Period | Order in units: | Price $ | Workers hired | Workers dismissed | Marketing expense $ | Dividends % | Loans $ | Training expense $ | R\&D expense $ | Sales forecast next period $ | Net income forecast $ | Investment |
|:---:|:---:|:---:|:---:|:---:|:---:|:---:|:---:|:---:|:---:|:---:|:---:|:---:|
| 1.00 | 4,000.00 | 110.00 | 0.00 | 0.00 | 0.00 | 0.00 | 0.00 | 0.00 | 0.00 | 0.00 | 0.00 | 0.00 |
| 2.00 | 4,000.00 | 110.00 | 0.00 | 0.00 | 0.00 | 0.00 | 0.00 | 0.00 | 0.00 | 0.00 | 0.00 | 0.00 |
| 3.00 | 4,000.00 | 110.00 | 0.00 | 0.00 | 0.00 | 0.00 | 0.00 | 0.00 | 0.00 | 0.00 | 0.00 | 0.00 |
| 4.00 | 4,000.00 | 110.00 | 0.00 | 0.00 | 0.00 | 0.00 | 0.00 | 0.00 | 0.00 | 0.00 | 0.00 | 0.00 |
| 5.00 | 4,000.00 | 110.00 | 0.00 | 0.00 | 0.00 | 0.00 | 0.00 | 0.00 | 0.00 | 0.00 | 0.00 | 0.00 |
| 6.00 | 4,000.00 | 110.00 | 0.00 | 0.00 | 0.00 | 0.00 | 0.00 | 0.00 | 0.00 | 0.00 | 0.00 | 0.00 |
| 7.00 | 3,000.00 | 110.00 | 0.00 | 0.00 | 0.00 | 0.00 | 0.00 | 0.00 | 0.00 | 0.00 | 0.00 | 0.00 |
| 8.00 | 3,000.00 | 110.00 | 0.00 | 0.00 | 0.00 | 0.00 | 0.00 | 0.00 | 0.00 | 0.00 | 0.00 | 0.00 |
| 9.00 | 3,000.00 | 110.00 | 0.00 | 0.00 | 0.00 | 0.00 | 0.00 | 0.00 | 0.00 | 0.00 | 0.00 | 0.00 |
| 10.00 | 3,000.00 | 110.00 | 0.00 | 0.00 | 0.00 | 0.00 | 0.00 | 0.00 | 0.00 | 0.00 | 0.00 | 0.00 |
| 11.00 | 3,000.00 | 110.00 | 0.00 | 0.00 | 0.00 | 0.00 | 0.00 | 0.00 | 0.00 | 0.00 | 0.00 | 0.00 |

| 12.00 | 3,000.00 | 110.00 | 0.00 | 0.00 | 0.00 | 0.00 | 0.00 | 0.00 | 0.00 | 0.00 | 0.00 | 0.00 |

Based on this information, please make the following decisions for January. Your response should be in a row as follows:

Your order in units (required) Price $ (required) Workers hired Workers dismissed Marketing expense $ Dividends % profit (enter as percentage) Loans $ Training expense $ R&D expense $ Sales forecast next period $ Net income forecast $

Format the decisions in format to paste in Excel so each value will be in a separate cell.

# Appendix B Prompts after the first period

**Subsequent Prompts Repeated Until Last Period (it should be followed by results of the previous month copied from the simulator spreadsheet)**

You are still the CEO. Your goal remains to maximize profit, market share, and sustainability.

Here are the results from the previous period, which reflect the decisions you made last month. Please use this new information to make the best decisions for the next month in the same format as above: